\documentclass[11pt]{article}
\PassOptionsToPackage{table}{xcolor}
\usepackage[preprint]{acl}
\usepackage{times}
\usepackage{latexsym}
\usepackage[T1]{fontenc}
\usepackage[utf8]{inputenc}
\usepackage{microtype}
\usepackage{amsmath,amssymb,bm,booktabs,graphicx}
\usepackage[table]{xcolor}
\usepackage{algorithm,algorithmic}
\usepackage{placeins}
\usepackage{multirow}
\definecolor{zomplight}{HTML}{E6F2E8}

\newcommand{\std}[1]{\raisebox{0.4pt}{\fontsize{5.5}{6}\selectfont$\pm$#1}}
\newcommand{\tablestyle}{\fontsize{8}{9.6}\selectfont\renewcommand{\arraystretch}{1.12}\setlength{\tabcolsep}{0.45pt}}
\newcommand{\densetablestyle}{\tablestyle\fontsize{7.5}{9.6}\selectfont}

\title{ZOMP: Zeroth-Order Multimodal Prompt Tuning\\for Vision--Language Models}
\author{
Sajjad Ghiasvand$^{1}$ \quad Yifan Yang$^{2}$ \quad Mahnoosh Alizadeh$^{1}$ \quad Ramtin Pedarsani$^{1}$ \\
$^{1}$Electrical and Computer Engineering \quad $^{2}$Computer Science \\
University of California, Santa Barbara \\
\texttt{\{sajjad,yifanyang,alizadeh,ramtin\}@ucsb.edu}
}
\hypersetup{
  pdftitle={ZOMP: Zeroth-Order Multimodal Prompt Tuning for Vision-Language Models},
  pdfauthor={Sajjad Ghiasvand, Yifan Yang, Mahnoosh Alizadeh, Ramtin Pedarsani},
  pdfsubject={Zeroth-order multimodal prompt tuning for vision-language models},
  pdfkeywords={vision-language models, prompt tuning, zeroth-order optimization, image-text retrieval},
  pdflang={en-US}
}
\begin{document}
\maketitle
\begin{abstract}
Prompt tuning adapts vision--language models with few trainable parameters, but computing their gradients still requires backpropagation through frozen encoders. We introduce ZOMP, a forward-only method for deep multimodal prompt tuning under a limited loss-query budget. ZOMP combines shared prompt factors with a budget-indexed rank schedule and gradient-correction momentum. Early updates search a rank-one subspace; later updates activate additional components without discarding the existing optimizer state. With 5,000 training-loss queries, ZOMP achieves 68.2\% average 16-shot accuracy across 13 datasets, improving on ZIP by 6.0 points, alongside gains in base-to-new and distribution-shift generalization. On five image--text retrieval datasets, it outperforms four derivative-free baselines at both training sizes, improving average bidirectional Recall@1 over BPT-VLM by 2.63 points. An exploratory five-benchmark language diagnostic comparison examines caption discrimination beyond the retrieval galleries. Component, modality, depth, and budget studies characterize the method's benefits. ZOMP requires an internal-prompt inference interface and reduces allocated training memory relative to a matched-architecture first-order reference.
\end{abstract}

\section{Introduction}
Vision--language models such as CLIP~\citep{radford2021learning} place images and text in a shared embedding space. Their similarity function supports both recognizing a category from its name and retrieving an image from a sentence. Downstream adaptation can improve these capabilities without updating the entire backbone. Prompt tuning~\citep{COOP,cocoop} learns a small set of context vectors, and multimodal variants insert these vectors at several layers of both encoders~\citep{khattak2023maple}. This gives the adaptation process access to intermediate visual and linguistic representations while preserving the pretrained weights.

Parameter efficiency, however, does not imply gradient efficiency. Even when the encoders are frozen, gradients of deep prompts depend on their intermediate activations. Backpropagation must retain or recompute those activations, introducing a training-memory cost far larger than the prompt parameters themselves. A forward-only alternative estimates updates from scalar losses at perturbed prompt values. The relevant resource is then a budget of loss evaluations: each measurement must improve prompts spanning both modalities and multiple layers.

\begin{figure*}[t]
\centering\includegraphics[width=\textwidth]{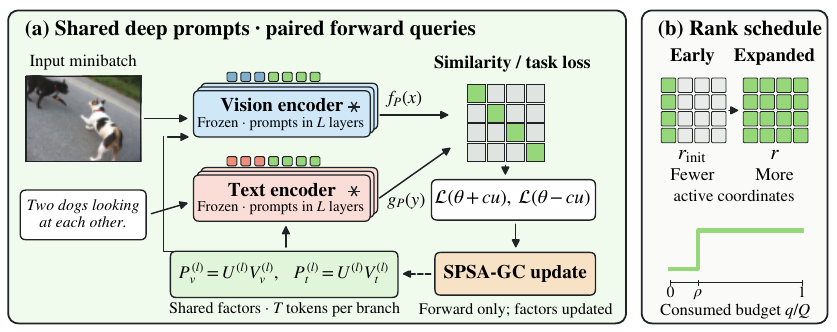}
\caption{\textbf{ZOMP overview.} Shared factors produce prompts at $L$ layers of each frozen encoder. Paired forward evaluations on the same minibatch estimate factor updates; SPSA-GC accumulates these estimates. The active rank grows from $r_{\rm init}$ to $r$, retaining learned factors and momentum. Blue and coral denote the image and text branches; green denotes trainable prompts. The grids illustrate restricted and expanded search spaces; the input pair and similarity matrix illustrate the computation.}
\label{fig:workflow}
\end{figure*}

This problem connects to black-box language-model adaptation, where compact prompt search spaces make derivative-free training practical~\citep{sun2022black,sun2022bbtv2}. In vision--language adaptation, BAR and BlackVIP learn visual input transformations~\citep{tsai2020transfer,oh2023blackvip}, ZIP optimizes a compact text-prompt representation~\citep{park2025zip}, and BPT-VLM learns prompts in both encoders~\citep{yu2023black}. These approaches establish that useful adaptation is possible without encoder gradients. They leave open how to allocate a finite budget across deep multimodal prompt coordinates.

The difficulty is that a two-sided perturbation provides a single directional loss difference, regardless of the number of coordinates it perturbs. Increasing prompt depth and adapting both modalities enlarge the search space without increasing the information in each pair of queries. Permanently restricting the prompt rank reduces this burden, but also restricts the functions available throughout training. We instead ask whether a restricted early search can be followed by a more expressive search within the same total budget.

We propose \textbf{ZOMP} (\textbf{Z}eroth-\textbf{O}rder \textbf{M}ultimodal \textbf{P}rompt tuning), shown in Figure~\ref{fig:workflow}. ZOMP introduces a budget-indexed rank schedule for forward-only optimization of shared deep multimodal prompts. The optimizer initially perturbs a subset of factor components, then activates further components as the budget is consumed. Gradient-correction momentum~\citep{oh2023blackvip} accumulates information from successive estimates, and the transition preserves both current prompt values and the momentum of already active coordinates. This procedure allocates early queries to a restricted search while retaining the capacity of the complete prompt for later refinement.

Our access model is explicit: inference must accept internal prompt embeddings and return embeddings or losses, but need not expose gradients. This includes local frozen-model execution and a prompt-aware inference service. It is a stronger interface than an API accepting only image pixels and text strings. We evaluate training-memory savings; arbitrary hosted APIs and edge deployment remain outside our scope.

We evaluate both recognition and sentence-conditioned image--text matching. The classification study covers 13 datasets, base-to-new generalization, cross-dataset transfer, and distribution shift. Retrieval evaluates COCO, Flickr8k, RSICD, Flickr30k, and TextCaps at two training-image counts with the same 5,000-query budget. ZOMP leads the compared derivative-free methods in all ten retrieval settings. Caption diagnostics then test whether the same retrieval-trained prompts distinguish semantic changes without diagnostic-specific training. These tests complement retrieval, where many incorrect candidates differ from a reference in several obvious ways.

\begin{figure*}[t]
\centering\includegraphics[width=\textwidth]{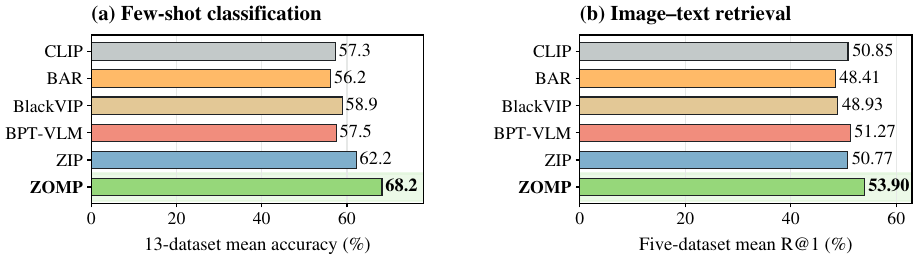}
\caption{\textbf{Classification and retrieval overview.} Left: mean 16-shot accuracy on all 13 datasets. Right: mean bidirectional R@1 over all five retrieval datasets and both training-image counts. Adapted methods use a 5,000-query cap; CLIP is unprompted. Tables~\ref{tab:fewshot} and~\ref{tab:retrieval} give per-dataset results and seed variability.}
\label{fig:overview}
\end{figure*}

Our contributions are threefold: \textbf{(1)} a budget-indexed rank schedule for forward-only optimization of shared deep multimodal prompts; \textbf{(2)} a broad empirical comparison on classification and five-dataset retrieval under explicit query and terminal-checkpoint protocols; and \textbf{(3)} controlled studies of components, modality, depth, query budget, and backbone size. Figure~\ref{fig:overview} summarizes classification and retrieval performance; ablations isolate scheduling, while language diagnostics provide exploratory evidence on transfer.

\section{Preliminaries}
\label{sec:prelim}
\paragraph{Vision--language matching.}
Let $f$ and $g$ denote frozen CLIP image and text encoders. Their normalized output embeddings define
\begin{equation}
 s(x,y)=\frac{\langle f(x),g(y)\rangle}{\tau},
 \label{eq:similarity}
\end{equation}
where $\tau$ is the pretrained temperature. Classification uses one text description $y_c$ per class and predicts with a softmax over $s(x,y_c)$. Retrieval instead compares complete captions with images in a gallery. Both tasks therefore adapt the same similarity function, while differing in their supervision and candidate sets.

\paragraph{Deep multimodal prompts.}
Prompt tuning learns context vectors while holding encoder weights fixed~\citep{COOP,cocoop}. Independent deep multimodal prompting~\citep{khattak2023maple} supplies separate matrices $P_v^{(l)}\in\mathbb R^{T\times d_v}$ and $P_t^{(l)}\in\mathbb R^{T\times d_t}$ at each of the first $L$ layers. Here $T$ is the number of prompt tokens and $d_v,d_t$ are the hidden widths, which need not equal the shared output dimension. The visual prompts accompany patch tokens; textual prompts occupy reserved context positions. Fresh prompts replace the prompt positions at deeper prompted layers, while image and caption content states continue through the transformer.

Direct optimization of these matrices uses $LT(d_v+d_t)$ trainable coordinates. Although this is small relative to the backbone, the gradient still traverses subsequent frozen layers. Each derivative-free update estimates all active coordinates from shared loss differences. Factorized prompts let ZOMP vary the search dimension by activating rank components.

\paragraph{Loss queries and SPSA.}
Collect trainable parameters in $\theta$ and let $\mathcal L(\theta;\mathcal B)$ denote the scalar loss on a minibatch. One \emph{training query} evaluates this loss over the full minibatch. Simultaneous perturbation stochastic approximation (SPSA)~\citep{SPSA} uses symmetric evaluations:
\begin{equation}
 \widehat g=\frac{1}{N}\sum_{i=1}^{N}
 \frac{\mathcal L(\theta+c z_i;\mathcal B)-\mathcal L(\theta-c z_i;\mathcal B)}{2c}\,z_i^{-1}.
 \label{eq:spsa}
\end{equation}
The reciprocal is elementwise and $c$ controls the perturbation radius. Independent directions $z_i$ have zero-mean coordinates with finite inverse moments. The estimate uses $2N$ loss queries independently of parameter dimension; its noise, however, depends on how many coordinates share each directional measurement~\citep{SPSA,ghadimi2013stochastic}. A budget $Q$ consequently controls the number of measurements, not the number of training images or epochs.

\paragraph{Relation to previous methods.}
BAR learns a visual input frame, and BlackVIP generates an image-dependent visual prompt using a coordinator and a frozen auxiliary encoder~\citep{tsai2020transfer,oh2023blackvip}. Both act in pixel space and can use an input-only interface. BlackVIP also introduces the SPSA-GC update adopted below. BPT-VLM uses evolutionary optimization for multimodal prompt tuning~\citep{yu2023black}. ZIP combines a low-rank text-prompt representation with a diagonal scaling matrix, an intrinsic-dimensional projection, and gradient clipping~\citep{park2025zip}. Shared low-rank factors can couple visual and textual prompts~\citep{ghiasvand2026mmlop}. ZOMP varies the active rank of these deep multimodal prompts over the query budget.

EffiBlackVIP learns visual-prompt subspaces through pretraining and subspace learning~\citep{liu2025query}. AdvCLIP-LoRA uses adversarial low-rank weight adaptation~\citep{ghiasvand2025adversarial}, while pFedMMA combines local modality-specific adapters with a globally shared projection for personalized federated learning~\citep{ghiasvand2026pfedmma}. ZOMP studies forward-only prompt adaptation under a fixed query budget. Matched controls vary modality, depth, rank scheduling, and optimizer; a first-order reference compares quality and memory using the same prompt architecture.

\section{Proposed Algorithm}
\label{sec:method}
ZOMP optimizes deep multimodal prompts under a fixed loss-query budget through shared factors, a budget-indexed rank schedule, and gradient-correction momentum. Section~\ref{sec:setup} specifies depth, token count, and schedule settings.

\subsection{Shared Multimodal Prompt Factors}
At layer $l$, we parameterize the prompt matrices as
\begin{equation}
 P_v^{(l)}=U^{(l)}V_v^{(l)},\qquad
 P_t^{(l)}=U^{(l)}V_t^{(l)},
 \label{eq:factor}
\end{equation}
where $U^{(l)}\in\mathbb R^{T\times r}$ is shared between modalities and $V_m^{(l)}\in\mathbb R^{r\times d_m}$ is modality-specific. The shared matrix controls how rank components combine across prompt positions. Updating it changes both branches, while updating a $V_m^{(l)}$ changes only its branch. Sharing is within each layer; different layers have different factors.

The rank components are explicit:
\begin{equation}
 P_m^{(l)}=\sum_{j=1}^{r}U_{:,j}^{(l)}V_{m,j,:}^{(l)}.
 \label{eq:components}
\end{equation}
Component $j$ comprises one column of $U^{(l)}$ and one row of each $V_m^{(l)}$. These coordinates are disjoint from those of other components, so a binary mask can freeze or activate a component without changing the optimizer or reallocating tensors. With $r_q$ active components, the dimension is
\begin{equation}
 D_q=Lr_q(T+d_v+d_t).
 \label{eq:dimension}
\end{equation}

Depth $L$ controls the adapted layers, token count $T$ sets the context positions, and rank controls their shared subspace. When $r=T$ and $U^{(l)}$ is invertible, any prompt pair can be represented by $V_m^{(l)}=(U^{(l)})^{-1}P_m^{(l)}$. At smaller rank, both branches share a restricted token-space subspace. Thus the framework can restrict the early search while retaining a more expressive family later; its final parameter count depends on the chosen rank.

We initialize $U^{(l)}\sim\mathcal N(0,\sigma_U^2)$ and $V_m^{(l)}=0$, with scale $\sigma_U>0$. Generated prompts initially vanish, but perturbing $V_m^{(l)}$ can affect the loss. Prompt positions still participate in attention, so this differs from unprompted CLIP.

\subsection{Budget-Indexed Rank Scheduling}
Let $q$ count consumed training queries and $Q$ denote the budget. A nondecreasing schedule $r(q)$ specifies the number of active components, up to the allocated rank $r$. A two-stage schedule is
\begin{equation}
 r(q)=\begin{cases}r_{\rm init},&q/Q\leq\rho,\\r,&q/Q>\rho,\end{cases}
 \label{eq:schedule}
\end{equation}
where $1\leq r_{\rm init}<r$ and $\rho\in(0,1)$ controls the expansion point. Additional stages can introduce intermediate ranks. At the start of each update, a mask $M_q$ activates the coordinates of the first $r(q)$ components at every prompted layer. Indexing the schedule by consumed budget makes its stages comparable across training horizons. Early queries refine a restricted component set; later queries explore the expanded family.

For each direction in Equation~\ref{eq:spsa}, we use $u_i=M_q\odot z_i$. Active entries of $z_i$ are independent draws from a symmetric distribution with finite second and inverse-second moments. The masked reciprocal equals $1/u_{i,j}$ for active coordinates and zero elsewhere, avoiding division by zero. Both factor perturbations and gradient estimates vanish on inactive coordinates. Their momentum is initialized to zero and remains zero until activation.

When a component becomes active, its row of $V_m^{(l)}$ is still zero. Its contribution in Equation~\ref{eq:components} therefore vanishes at the transition, preserving the current prompt values. Subsequent updates explore the new directions while retaining existing factors and their optimizer state.

\paragraph{Scheduling rationale.}
At a fixed query budget, rank controls how many factor coordinates each estimate must update. Starting with fewer active components concentrates the early search in a smaller prompt family; later activation permits adjustments outside that family while preserving the learned prompts at the transition. This motivates the schedule, whose empirical value is assessed against fixed-rank controls in Section~\ref{sec:results}.

\subsection{Gradient-Correction Updates}
ZOMP averages up to $N$ paired directions and applies the SPSA-GC update of \citet{oh2023blackvip}:
\begin{align}
 m_k&=\beta m_{k-1}+\widehat g_k,\\
 \theta_{k+1}&=\theta_k-\eta_k(\widehat g_k+\beta m_k),
 \label{eq:gc}
\end{align}
where $\beta\in[0,1)$ controls momentum. Positive gain sequences $c_k$ and $\eta_k$ set the perturbation radius and step size; we use power-law schedules $c_k=c/k^\gamma$ and $\eta_k=a/(1+k)^\alpha$. Each positive--negative pair shares a minibatch and its preprocessing realization. Updates use $2N$ queries, or fewer at budget or evaluation boundaries.

The update combines current and accumulated estimates. As the mask expands, existing coordinates retain their momentum and newly active coordinates begin with zero history. Algorithm~\ref{alg:zomp} summarizes training. Ablations compare fixed and scheduled ranks, with and without momentum.

\begin{algorithm}[t]
\caption{ZOMP under a loss-query budget}
\label{alg:zomp}
\small
\begin{algorithmic}[1]
\REQUIRE Frozen encoders, data $\mathcal D$, even budget $Q$, schedule $r(q)$, gains $\{c_k,\eta_k\}$, $N$ probes, momentum $\beta$
\STATE Initialize factors as in Section~\ref{sec:method}; $m=0$, $q=0$, $k=1$
\WHILE{$q<Q$}
\STATE Sample one minibatch $\mathcal B$; form $M_q$ from the active components specified by $r(q)$
\STATE Set $n=\min(N,(Q-q)/2)$, shortened at an evaluation boundary if needed
\STATE Compute $\widehat g_k$ from $n$ paired probes using Equation~\ref{eq:spsa}, $u_i=M_q\odot z_i$, and the masked reciprocal
\STATE Update $m_k$ and $\theta_{k+1}$ using Equation~\ref{eq:gc}
\STATE $q\leftarrow q+2n$; $k\leftarrow k+1$
\ENDWHILE
\RETURN terminal factors $\theta$
\end{algorithmic}
\end{algorithm}

\subsection{Objectives and Inference}
Classification minimizes cross-entropy over class-name descriptions. For retrieval, a batch contains $B$ distinct images and one sampled reference caption per image. Writing $s_{ij}=s_\theta(x_i,y_j)$ for prompted similarities, the loss is
\begin{equation}
\mathcal L_{\rm ret}=-\frac{1}{2B}\sum_i\left(
\log\frac{e^{s_{ii}}}{\sum_j e^{s_{ij}}}
+\log\frac{e^{s_{ii}}}{\sum_j e^{s_{ji}}}\right).
\label{eq:retrieval}
\end{equation}
The two terms learn image-to-text and text-to-image discrimination using other batch elements as negatives. The encoders and temperature remain frozen. Text prompt positions precede the caption, preserving its words subject to CLIP's context limit. Thus the retrieval objective changes the supervision, without requiring a different rank schedule or a diagnostic-specific prompt.

At inference, the terminal factors generate fixed layerwise prompts. Image and caption embeddings can be computed independently and cached; classification and retrieval use their similarities. No perturbations, momentum, or loss evaluations are needed to rank a gallery. The same prompts are transferred to caption diagnostics without further optimization. Training therefore requires an internal-prompt forward interface, while the optimizer itself observes only scalar losses and never differentiates through the encoders.

\section{Empirical Results}
\label{sec:results}
\begin{table*}[t]
\caption{Few-shot performance on 13 datasets (16-shot, 5,000 queries, ViT-B/16). Values are mean\,$\pm$\,SD over three seeds; Manual Prompt is unprompted CLIP. \textbf{Best} and \underline{second-best} adapted methods are marked.}
\label{tab:fewshot}\label{tab:classification}\label{tab:broad_fewshot}
\centering\densetablestyle
\begin{tabular*}{\linewidth}{@{\extracolsep{\fill}}lccccccccccccc>{\columncolor[gray]{0.9}}c@{}}
\toprule
\multicolumn{1}{c}{Method} & \rotatebox[origin=c]{90}{Caltech101} & \rotatebox[origin=c]{90}{OxfordPets} & \rotatebox[origin=c]{90}{Flowers102} & \rotatebox[origin=c]{90}{Food101} & \rotatebox[origin=c]{90}{FGVCAircraft} & \rotatebox[origin=c]{90}{SUN397} & \rotatebox[origin=c]{90}{DTD} & \rotatebox[origin=c]{90}{SVHN} & \rotatebox[origin=c]{90}{EuroSAT} & \rotatebox[origin=c]{90}{RESISC45} & \rotatebox[origin=c]{90}{CLEVR} & \rotatebox[origin=c]{90}{UCF101} & \rotatebox[origin=c]{90}{ImageNet} & \rotatebox[origin=c]{90}{\textbf{Average}} \\
\midrule
Manual Prompt & 93.2 & 89.1 & 70.8 & 85.9 & 24.8 & 62.6 & 44.1 & 19.2 & 48.4 & 57.2 & 15.2 & 67.5 & 66.7 & 57.3 \\
BAR & 91.8\std{0.2} & 87.8\std{0.1} & \underline{69.9}\std{0.3} & 83.7\std{0.1} & 21.7\std{0.2} & 60.8\std{0.1} & 40.0\std{0.1} & 19.7\std{0.7} & 61.6\std{0.2} & 49.9\std{0.4} & 16.0\std{0.1} & 64.0\std{0.1} & 63.5\std{0.2} & 56.2\std{0.2} \\
\textsc{BlackVIP} & 92.5\std{0.1} & 88.4\std{0.3} & 69.4\std{0.7} & 84.0\std{0.1} & 23.6\std{0.2} & 61.9\std{0.3} & 43.1\std{0.2} & 30.2\std{3.4} & 54.7\std{5.1} & 57.1\std{0.4} & \underline{29.5}\std{3.0} & 66.4\std{0.2} & 64.6\std{0.3} & 58.9\std{0.7} \\
BPT-VLM & 89.1\std{0.6} & 89.6\std{0.7} & 66.9\std{2.7} & 84.0\std{0.4} & 24.0\std{1.7} & 53.5\std{2.1} & 44.7\std{1.4} & 33.0\std{0.9} & 61.1\std{5.0} & 59.9\std{1.5} & 19.9\std{5.6} & 63.6\std{1.2} & 58.7\std{1.2} & 57.5\std{0.8} \\
\textsc{ZIP} & \underline{93.1}\std{0.8} & \underline{91.9}\std{0.4} & 69.3\std{1.2} & \underline{86.1}\std{0.3} & \underline{26.0}\std{0.4} & \underline{62.2}\std{0.8} & \underline{47.8}\std{2.3} & \underline{43.3}\std{2.4} & \underline{64.6}\std{3.5} & \underline{65.5}\std{0.5} & 24.8\std{3.0} & \underline{67.9}\std{1.2} & \underline{65.8}\std{1.5} & \underline{62.2}\std{0.3} \\
\textbf{ZOMP} & \textbf{94.8}\std{0.3} & \textbf{93.0}\std{0.2} & \textbf{76.5}\std{1.3} & \textbf{87.1}\std{0.1} & \textbf{28.2}\std{0.5} & \textbf{68.4}\std{0.7} & \textbf{56.4}\std{1.3} & \textbf{51.4}\std{6.1} & \textbf{80.4}\std{1.8} & \textbf{73.5}\std{0.5} & \textbf{35.1}\std{2.3} & \textbf{73.6}\std{1.5} & \textbf{68.7}\std{0.2} & \textbf{68.2}\std{0.4} \\
\bottomrule
\end{tabular*}

\end{table*}

\subsection{Experimental Setup}
\label{sec:setup}
\paragraph{Tasks and comparisons.}
We retain the 13-dataset classification suite and evaluate image--text retrieval on COCO~\citep{karpathy2015deep}, Flickr8k~\citep{hodosh2013framing}, RSICD~\citep{lu2018exploring}, Flickr30k~\citep{young2014flickr30k}, and TextCaps~\citep{sidorov2020textcaps}. Classification uses 16 examples per class; retrieval uses $n_{\rm train}\in\{1{,}024,4{,}096\}$ images, with nested subsets per seed. These are \emph{training-image counts}, not query budgets. Models are tested on their held-out galleries and caption diagnostics.

We compare BAR, BlackVIP, BPT-VLM, and ZIP. Retrieval uses Equation~\ref{eq:retrieval} for every adapted method. A separate AdamW reference trains the same deep factorized prompts with exact gradients for 20 epochs. Appendix~\ref{app:protocol} specifies baseline adaptations and the broad versus controlled classification protocols.

\paragraph{Prompt and optimizer settings.}
Unless ablated, ZOMP inserts $T=4$ prompt tokens in the first $L=9$ layers of both encoders. We set $r_{\rm init}=1$, final rank $r=4$, and expansion fraction $\rho=0.2$. Each update averages up to $N=5$ paired probes with momentum $\beta=0.8$ and no gradient clipping. Initialization, perturbation sampling, and gain schedules are specified in Appendix~\ref{app:protocol}; Section~\ref{sec:ablations} reports their ablations.

\paragraph{Budget and reporting.}
Unless stated otherwise, the backbone is CLIP ViT-B/16 and the training cap is $Q=5{,}000$ scalar minibatch loss queries. Retrieval and the controlled studies use batch size 128. Every batch evaluated by a BPT candidate counts toward the cap. Validation and final inference are recorded separately. Hyperparameters are developed on validation data; primary scores use terminal checkpoints for all methods. We report means and sample SDs over three seeds, computing dataset averages within seed first. Appendix~\ref{app:data} details splits, preprocessing, and development.

\subsection{Few-Shot Classification and Generalization}
Figure~\ref{fig:overview} summarizes the classification and retrieval comparisons.
Table~\ref{tab:fewshot} reports all 13 classification datasets. ZOMP reaches \textbf{68.2\%} average accuracy, compared with ZIP's 62.2\%, and leads the compared adapted methods on every dataset. Improvements span fine-grained recognition, texture, and remote sensing: relative to ZIP, Flowers102 increases from 69.3 to 76.5, DTD from 47.8 to 56.4, and EuroSAT from 64.6 to 80.4. These results establish the complete method's performance; the component study below isolates the smaller contribution of rank scheduling.

\begin{table*}[t]
\caption{Base-to-new generalization: per-dataset harmonic mean (HM) of base and new accuracy. All methods use 16 shots and a 5,000-query cap; mean\,$\pm$\,SD over three seeds. Average is the mean of dataset HMs. Base and new accuracies are provided separately in Appendix~\ref{app:classification}. \textbf{Best} and \underline{second-best} adapted methods are marked.}
\label{tab:base2new}\label{tab:broad_hm}
\centering\densetablestyle
\begin{tabular*}{\linewidth}{@{\extracolsep{\fill}}lccccccccccccc>{\columncolor[gray]{0.9}}c@{}}
\toprule
\multicolumn{1}{c}{Method} & \rotatebox[origin=c]{90}{Caltech101} & \rotatebox[origin=c]{90}{OxfordPets} & \rotatebox[origin=c]{90}{Flowers102} & \rotatebox[origin=c]{90}{Food101} & \rotatebox[origin=c]{90}{FGVCAircraft} & \rotatebox[origin=c]{90}{SUN397} & \rotatebox[origin=c]{90}{DTD} & \rotatebox[origin=c]{90}{SVHN} & \rotatebox[origin=c]{90}{EuroSAT} & \rotatebox[origin=c]{90}{RESISC45} & \rotatebox[origin=c]{90}{CLEVR} & \rotatebox[origin=c]{90}{UCF101} & \rotatebox[origin=c]{90}{ImageNet} & \rotatebox[origin=c]{90}{Average} \\
\midrule
BAR & 94.8\std{0.0} & 92.7\std{0.0} & \underline{73.7}\std{0.2} & 88.8\std{0.0} & 27.4\std{0.1} & 70.8\std{0.1} & 51.0\std{1.1} & 25.8\std{0.7} & \textbf{76.6}\std{0.2} & 62.1\std{0.1} & 30.5\std{0.0} & 71.5\std{0.1} & 67.4\std{0.1} & 64.1\std{0.1} \\
\textsc{BlackVIP} & 94.7\std{0.1} & 93.1\std{0.1} & \underline{73.7}\std{0.6} & 89.1\std{0.2} & 29.8\std{0.4} & \underline{71.4}\std{0.4} & 55.0\std{0.4} & 36.2\std{6.7} & 49.1\std{1.6} & 66.3\std{0.7} & \underline{39.6}\std{2.9} & \underline{73.8}\std{0.6} & 68.4\std{0.0} & 64.6\std{0.5} \\
BPT-VLM & 93.3\std{1.1} & 93.6\std{1.0} & 70.7\std{2.0} & 89.0\std{0.3} & \underline{31.5}\std{1.3} & 63.3\std{3.5} & 51.4\std{0.6} & 43.0\std{1.3} & \underline{74.6}\std{4.8} & 67.1\std{1.0} & 31.0\std{2.4} & 69.6\std{2.3} & 62.3\std{1.3} & 64.6\std{0.8} \\
\textsc{ZIP} & \underline{95.7}\std{0.5} & \underline{95.9}\std{0.3} & 72.4\std{0.3} & \underline{90.2}\std{0.5} & 30.4\std{1.1} & 70.4\std{1.4} & \underline{56.1}\std{1.4} & \underline{49.7}\std{5.0} & 69.8\std{5.5} & \underline{72.0}\std{0.8} & 33.9\std{1.3} & 72.6\std{2.6} & \underline{68.7}\std{1.3} & \underline{67.5}\std{0.7} \\
\textbf{ZOMP} & \textbf{95.8}\std{0.5} & \textbf{96.4}\std{0.2} & \textbf{79.3}\std{0.9} & \textbf{91.2}\std{0.1} & \textbf{33.1}\std{0.5} & \textbf{76.2}\std{0.4} & \textbf{61.1}\std{2.3} & \textbf{51.1}\std{3.3} & 71.9\std{7.4} & \textbf{77.1}\std{1.1} & \textbf{43.9}\std{2.8} & \textbf{77.1}\std{0.7} & \textbf{71.8}\std{0.2} & \textbf{71.2}\std{0.3} \\
\bottomrule
\end{tabular*}

\end{table*}

\paragraph{Base-to-new generalization.}
ZOMP obtains 76.3\% base accuracy and 67.6\% new accuracy, giving a 71.2\% average HM versus ZIP's 67.5\% (Table~\ref{tab:base2new}). Its new-class average also exceeds ZIP's 64.4\%, so the gain is not confined to fitted classes. HM improvements are particularly clear on Flowers102 (79.3 versus 72.4), SUN397 (76.2 versus 70.4), and CLEVR (43.9 versus 33.9).

\begin{table}[t]
\caption{Distribution shift of the ImageNet-trained prompt at 5,000 queries. Source is excluded from the four-target Average. Values are mean\,$\pm$\,SD; \textbf{best} and \underline{second-best} per column.}
\label{tab:ood}\label{tab:broad_ood}
\centering\tablestyle
\begin{tabular*}{\linewidth}{@{\extracolsep{\fill}}lccccc>{\columncolor[gray]{0.9}}c@{}}
\toprule
\multicolumn{1}{c}{Method} & \rotatebox[origin=c]{90}{Source} & \rotatebox[origin=c]{90}{ImageNet-A} & \rotatebox[origin=c]{90}{ImageNetV2} & \rotatebox[origin=c]{90}{ImageNet-R} & \rotatebox[origin=c]{90}{ImageNet-Sketch} & \rotatebox[origin=c]{90}{\textbf{Average}} \\
\midrule
BAR & 63.5\std{0.2} & 40.0\std{0.2} & 57.2\std{0.3} & 71.9\std{0.1} & 43.7\std{0.1} & 53.2\std{0.1} \\
\textsc{BlackVIP} & 64.6\std{0.3} & 39.1\std{0.3} & 58.0\std{0.1} & 72.0\std{0.2} & 43.6\std{0.1} & 53.2\std{0.1} \\
BPT-VLM & 58.7\std{1.2} & 33.9\std{4.6} & 45.9\std{3.5} & 62.9\std{3.7} & 31.9\std{1.4} & 43.7\std{3.3} \\
\textsc{ZIP} & \underline{65.8}\std{1.5} & \underline{47.8}\std{1.0} & \underline{59.5}\std{1.5} & \underline{75.1}\std{2.0} & \underline{45.4}\std{1.2} & \underline{57.0}\std{1.3} \\
\textbf{ZOMP} & \textbf{68.6}\std{0.2} & \textbf{50.1}\std{0.1} & \textbf{62.3}\std{0.3} & \textbf{76.6}\std{0.1} & \textbf{47.5}\std{0.2} & \textbf{59.1}\std{0.1} \\
\bottomrule
\end{tabular*}

\end{table}
\paragraph{Transfer and distribution shift.}
On four ImageNet shifts, ZOMP achieves the highest score in each column and a 59.1 target average, compared with ZIP's 57.0 (Table~\ref{tab:ood}). Transferring the same source prompt to 12 other datasets without target adaptation gives 56.9, compared with 55.5 for BAR, the strongest adapted baseline on average, and 53.8 for ZIP. This evaluates whether source-trained prompts generalize to new recognition tasks. Appendix~\ref{app:classification} gives per-target results.

\begin{table*}[t]
\caption{Retrieval: mean bidirectional R@1 (\%), mean\,$\pm$\,SD. Subcolumns count \textbf{training images}; all adapted methods use a \textbf{5,000-query cap}. Average equally weights five datasets. $\dagger$: modified galleries (Appendix~\ref{app:data}). \textbf{Best} and \underline{second-best} compare adapted methods.}
\label{tab:retrieval}
\centering\tablestyle
\begin{tabular*}{\linewidth}{@{\extracolsep{\fill}}l*{10}{c}>{\columncolor[gray]{0.9}}c>{\columncolor[gray]{0.9}}c@{}}
\toprule
 & \multicolumn{2}{c}{COCO} & \multicolumn{2}{c}{Flickr8k} & \multicolumn{2}{c}{RSICD} & \multicolumn{2}{c}{Flickr30k$^{\dagger}$} & \multicolumn{2}{c}{TextCaps$^{\dagger}$} & \multicolumn{2}{c}{\cellcolor[gray]{0.9}\textbf{Average}}\\
\cmidrule(lr){2-3}\cmidrule(lr){4-5}\cmidrule(lr){6-7}\cmidrule(lr){8-9}\cmidrule(lr){10-11}\cmidrule(lr){12-13}
\multicolumn{1}{c}{Method} & 1,024 & 4,096 & 1,024 & 4,096 & 1,024 & 4,096 & 1,024 & 4,096 & 1,024 & 4,096 & 1,024 & 4,096 \\
\midrule
Manual Prompt & 42.76 & 42.76 & 64.66 & 64.66 & 6.21 & 6.21 & 74.43 & 74.43 & 66.20 & 66.20 & 50.85 & 50.85 \\
BAR & 39.96\std{0.20} & 40.23\std{0.04} & 62.64\std{0.32} & 62.12\std{0.11} & 4.62\std{0.13} & 4.68\std{0.06} & 72.64\std{0.26} & 72.68\std{0.20} & 62.24\std{0.17} & 62.31\std{0.06} & 48.42\std{0.03} & 48.40\std{0.03} \\
\textsc{BlackVIP} & 41.19\std{0.43} & 41.32\std{0.14} & 62.99\std{0.40} & 62.62\std{0.31} & 4.58\std{0.17} & 5.03\std{0.28} & 73.78\std{0.58} & 73.49\std{0.52} & 61.96\std{0.34} & 62.35\std{0.36} & 48.90\std{0.13} & 48.96\std{0.01} \\
BPT-VLM & \underline{44.06}\std{0.19} & 43.76\std{0.25} & \underline{65.27}\std{0.73} & \underline{65.37}\std{0.36} & 5.67\std{0.57} & 5.68\std{0.22} & \underline{74.97}\std{0.20} & \underline{74.96}\std{0.85} & \underline{66.63}\std{0.52} & \underline{66.34}\std{0.09} & \underline{51.32}\std{0.13} & \underline{51.22}\std{0.16} \\
\textsc{ZIP} & 42.41\std{0.84} & \underline{43.90}\std{0.71} & 64.42\std{0.86} & 64.32\std{0.72} & \underline{5.89}\std{0.24} & \underline{5.89}\std{0.41} & 74.58\std{0.13} & 74.78\std{1.30} & 65.82\std{0.52} & 65.66\std{0.65} & 50.62\std{0.05} & 50.91\std{0.43} \\
\textbf{ZOMP} & \textbf{46.53}\std{0.10} & \textbf{46.37}\std{0.19} & \textbf{68.67}\std{0.26} & \textbf{68.74}\std{0.66} & \textbf{6.92}\std{0.52} & \textbf{6.94}\std{0.13} & \textbf{78.03}\std{0.44} & \textbf{78.65}\std{0.50} & \textbf{68.94}\std{0.12} & \textbf{69.20}\std{0.28} & \textbf{53.82}\std{0.06} & \textbf{53.98}\std{0.28} \\
\bottomrule
\end{tabular*}

\end{table*}

\begin{figure*}[t]
\centering\includegraphics[width=\textwidth]{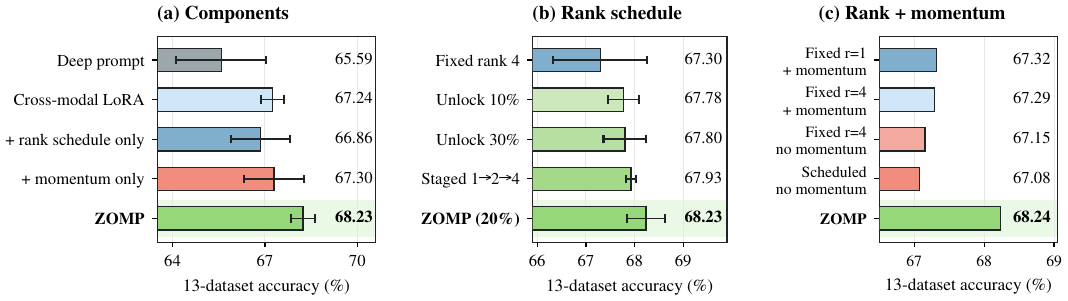}
\caption{\textbf{Components, rank schedules, and matched optimizer controls.} All panels average 13 classification datasets at 5,000 queries with $(a,c)=(0.01,0.01)$. (a) Original components: each ``+'' independently modifies Cross-modal LoRA. (b) Schedule sensitivity with SPSA-GC. (c) New rank/momentum controls without clipping. Panels (a,b) show three-seed mean\,$\pm$\,SD; panel (c) uses seed 1 on ImageNet and three seeds elsewhere, without error bars. Tables~\ref{tab:broad_components}, \ref{tab:broad_sensitivity}, and~\ref{tab:class_controls} give full results.}
\label{fig:ablations}\label{fig:native_ablation}
\end{figure*}

\subsection{Image--Text Retrieval}
\label{sec:retrieval}
Table~\ref{tab:retrieval} extends evaluation from class names to complete captions. ZOMP leads all four adapted baselines in all ten dataset--size settings. Its equal-setting mean bidirectional R@1 is \textbf{53.90}, versus 51.27 for BPT-VLM and 50.77 for ZIP: gains of 2.63 and 3.13 points. Each setting ranks its complete held-out gallery.

At 4,096 training images, ZOMP reaches 46.37 on COCO and 68.74 on Flickr8k. Paired seed/image-cluster gains over BPT-VLM are $+2.60$ (95\% interval $[2.11,3.09]$) and $+3.36$ ($[2.02,4.72]$), respectively. These intervals are descriptive and unadjusted. Appendix~\ref{app:retrieval} gives directional recalls and first-order references. Flickr30k uses a filtered 761-image gallery; TextCaps uses 3,166 validation images as its retrieval test gallery. These protocols differ from official evaluations.

\begin{table*}[t]
\caption{Caption-contrast diagnostics (\%): terminal models, five adaptation datasets, both training sizes, three seeds, 5,000-query cap. Average equally weights the five benchmarks. Benchmark definitions and scoring are given in Appendix~\ref{app:language}. $\dagger$: official EqBen 10\% subset, group score. \textbf{Best}/\underline{second-best} compare adapted methods.}
\label{tab:language}
\centering\tablestyle
\begin{tabular*}{\linewidth}{@{\extracolsep{\fill}}lccccc>{\columncolor[gray]{0.9}}c@{}}
\toprule
\multicolumn{1}{c}{Method} & SugarCrepe & CLEVR foils & SugarCrepe++ & What'sUp & EqBen$^{\dagger}$ & Average \\
\midrule
Manual Prompt & 73.09 & 58.93 & 59.71 & 42.95 & 10.29 & 48.99 \\
BAR & 71.95\std{0.02} & 58.56\std{0.03} & \underline{60.30}\std{0.03} & \textbf{44.05}\std{0.02} & 10.13\std{0.03} & \underline{49.00}\std{0.01} \\
\textsc{BlackVIP} & 72.67\std{0.26} & 57.81\std{0.03} & 58.88\std{0.29} & \underline{43.80}\std{0.28} & 9.69\std{0.17} & 48.57\std{0.05} \\
BPT-VLM & \textbf{75.79}\std{0.19} & 58.52\std{0.57} & 58.04\std{0.53} & 41.68\std{0.23} & \underline{10.91}\std{0.06} & 48.99\std{0.22} \\
\textsc{ZIP} & 74.88\std{0.11} & \textbf{59.27}\std{0.05} & 57.77\std{0.80} & 41.86\std{0.06} & 10.29\std{0.17} & 48.81\std{0.17} \\
\textbf{ZOMP} & \underline{75.27}\std{0.29} & \underline{59.13}\std{0.44} & \textbf{61.93}\std{0.24} & 42.08\std{0.34} & \textbf{11.03}\std{0.16} & \textbf{49.89}\std{0.23} \\
\bottomrule
\end{tabular*}

\end{table*}

\subsection{Language Diagnostics}
\label{sec:language}
We apply retrieval-trained prompts without additional optimization to SugarCrepe~\citep{hsieh2023sugarcrepe}, scene-verified CLEVR foils~\citep{CLEVR}, SugarCrepe++~\citep{dumpala2024sugarcrepepp}, What'sUp~\citep{kamath2023whatsup}, and EqBen~\citep{wang2023eqben}. SugarCrepe tests caption edits; SugarCrepe++ requires two semantically equivalent positive captions to outrank a negative. CLEVR foils test counting, color binding, and relation roles. What'sUp tests spatial caption choice, while EqBen requires consistent matching of two images and two captions. Appendix~\ref{app:language} details scoring and overlap filtering for all retained categories of each selected benchmark.

ZOMP averages \textbf{49.89}, versus 49.00 for BAR and 48.99 for BPT-VLM: a \textbf{0.89-point} lead (Table~\ref{tab:language}). Its largest gain is on SugarCrepe++, at 61.93 versus BAR's 60.30. ZOMP leads EqBen by 0.12 points over BPT-VLM; other methods lead SugarCrepe, CLEVR, and What'sUp. This selected average is exploratory. Diagnostic labels never determine training, hyperparameters, or checkpoints. Appendix~\ref{app:language_results} groups scores by training data.

\subsection{Ablation Studies}
\label{sec:ablations}
\paragraph{Components and scheduling.}
Figure~\ref{fig:ablations}a shows the original components across all 13 datasets. Direct deep prompts with SPSA average 65.59, the clipped factorized configuration 67.24, and ZOMP 68.23. Within SPSA-GC, rank scheduling improves 67.30 to 68.23, a 0.94-point gain. Most of the complete method's advantage over ZIP is already present before this increment. Appendix~\ref{app:ablation} gives per-dataset results and clipping settings.

Figure~\ref{fig:native_ablation}c compares fixed rank one and unclipped optimizer controls under the original protocol across all 13 datasets: three seeds per dataset, except a common seed 1 on ImageNet. ZOMP averages 68.24, versus 67.32 for fixed rank one and 67.29 for fixed rank four with momentum. Without momentum, fixed rank four and scheduling reach 67.15 and 67.08. Thus the complete method leads these controls, with scheduling providing a 0.95-point gain within SPSA-GC. Scheduling has smaller effects in the initial retrieval controls.

Fixed rank one tests whether restricting the search alone suffices; ZOMP improves its mean by 0.92 points. On the 12 datasets with three seeds, ZOMP averages 68.20 versus 67.23 for fixed rank one, so this advantage also holds without the single-seed ImageNet result. All retained ZOMP runs reproduce the original test counts. Scheduling gains 0.95 points with momentum but changes the plain-SPSA mean by $-0.07$. Removing momentum from the scheduled model costs 1.16 points, compared with 0.14 at fixed rank four. These controls support the joint design at fixed gains; they do not compare separately tuned optimizers.

\begin{figure}[t]
\centering\includegraphics[width=\columnwidth]{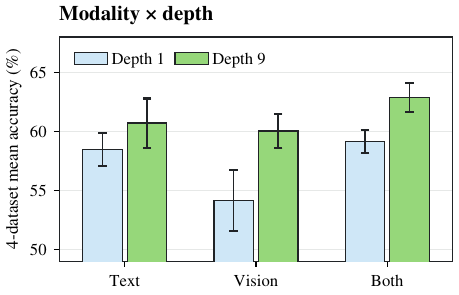}
\caption{\textbf{Modality and prompt depth.} Four-dataset classification mean with $(a,c)=(0.01,0.005)$ at 5,000 queries. Error bars give sample SD over three seeds. Table~\ref{tab:modality} reports per-dataset results and the separate COCO control.}
\label{fig:modality}
\end{figure}

\paragraph{Sensitivity, modality, and depth.}
Unlocking at 10\% or 30\% of the budget gives 67.78 or 67.80; a staged $1\!\to\!2\!\to\!4$ schedule gives 67.93, close to the default's 68.23 (Figure~\ref{fig:ablations}b). Unsharing the token-space factor gives 67.52. Figure~\ref{fig:modality} varies prompted branches and depth on the separate four-dataset suite: at depth nine, text-only, vision-only, and dual-encoder prompts reach 60.71, 60.05, and 62.88. At depth one, dual-encoder prompting reaches 59.16. These controls support deep multimodal prompting for this suite, while Appendix~\ref{app:controlled} retains the different COCO ordering.

\subsection{Budget, Backbone, and Training Memory}
ZOMP exceeds ZIP on held-out DTD and COCO performance at each of three independent budgets: 500, 2,000, and 5,000 queries (Appendix~\ref{app:resources}). Each point uses its own training horizon. An offline diagnostic (Appendix~\ref{app:gradient}) finds lower relative estimator error during rank-one search.

On COCO with ViT-B/16 and batch size 128, ZOMP uses 1.31\,GiB peak allocated training memory versus 15.52\,GiB for the matched-architecture AdamW reference. AdamW reaches 51.54 R@1 after its 20-epoch horizon, versus 46.37 for ZOMP. The 11.8-fold memory reduction therefore accompanies a 5.17-point recall gap. Avoiding backward activation storage reduces training memory even though both methods update the same prompt family.  With ViT-L/14 at batch size 32, ZOMP reaches 48.19 COCO R@1 versus ZIP's 43.46, and 77.10 versus 74.05 on the DTD/Pets mean. Appendices~\ref{app:controlled} and~\ref{app:resources} give full results, including the original training curves and classification efficiency comparison.

\FloatBarrier
\section{Conclusion}
ZOMP optimizes shared deep multimodal prompts through forward-only loss evaluations, expanding the active rank as the query budget is consumed. It improves on the compared derivative-free methods across 13-dataset few-shot classification and all ten retrieval settings, with strong base-to-new and distribution-shift results. Focused ablations support scheduling within SPSA-GC on classification, while the exploratory language diagnostics characterize task-dependent transfer. The method supports adaptation through internal prompts when backpropagation is costly or unavailable.

\clearpage
\section*{Limitations}
\paragraph{Model and task coverage.}
Our evaluation focuses on English-language contrastive vision--language models and image--text similarity tasks. VISTA and MORFI~\citep{khezresmaeilzadeh2025vista,khezresmaeilzadeh2025morfi} explore training-free financial forecasting by combining stock-price sequences with line charts. Extending ZOMP to such generative tasks, multilingual encoders, and open-ended visual reasoning requires further study. Our benchmarks do not establish the same gains in these settings.

\paragraph{Access and computational resources.}
The framework assumes that frozen encoders accept internal prompt embeddings. Its applicability therefore depends on the interface exposed by a model. Forward-only optimization reduces training activation memory but requires repeated loss evaluations; its wall-clock cost depends on the backbone, query budget, and execution hardware. Resource trade-offs at substantially larger model scales remain to be evaluated.

\paragraph{Real-world generalization.}
Public benchmarks cover a finite range of visual domains and language constructions. Their results do not fully characterize robustness to evolving data, ambiguous descriptions, or unfamiliar deployment conditions. Broader evaluation is needed before using adapted models in applications where retrieval errors have consequential effects.

\section*{Ethical Considerations}
The experiments reuse established image--text and classification datasets without collecting new personal data or involving human participants. These datasets and CLIP can retain societal biases and sensitive content. Better retrieval can reproduce those biases more effectively; the diagnostics do not measure fairness or suitability for consequential decisions. Counterfactual audits of LLMs find demographic disparities in interaction quality even when refusal rates are uniform~\citep{amirimargavi2026equal}, motivating evaluation of distinct fairness dimensions. Pre-specified benchmarks for LLM advisors separately examine honesty under preference misalignment~\citep{hasanibalyani2026truthful}. Reproduction materials should include code, split identifiers, and derived measurements under the datasets' access conditions.

\bibliography{references}
\clearpage
\appendix
\section{Data and Evaluation Protocols}
\label{app:data}
\subsection{Retrieval Data}
\paragraph{COCO.}
We use the Karpathy image--caption split~\citep{karpathy2015deep}, sampling its training split without restval. Identifiable diagnostic images are excluded from adaptation and validation. The eligible training pool contains 82,150 images and the modified validation gallery 4,551 images. The test gallery retains 5,000 images and 25,010 captions: ten images have six references and the other 4,990 have five.

\paragraph{Flickr8k.}
The supplied package~\citep{hodosh2013framing} contains 6,000 training and 1,000 test images but lacks the official validation partition. We reserve 600 supplied training images for validation. Removing one exact duplicate leaves 5,399 eligible training images. The test gallery contains 1,000 images and 5,000 captions.

\paragraph{RSICD.}
The supplied data~\citep{lu2018exploring} contain 8,734 training, 1,094 validation, and 1,093 test images. Removing one exact duplicate leaves 8,733 eligible training images. Each image has five references. The test annotations contain 150 normalized caption strings shared by multiple images. Relevance follows the annotations; a shared caption does not make another image relevant.

\paragraph{Flickr30k.}
Starting from Karpathy-split metadata~\citep{young2014flickr30k,karpathy2015deep}, we exclude all supplied Flickr8k image identities from every split. A fixed hash-based selection defines a 6,000-image training pool; the remaining validation and test galleries contain 776 and 761 images. Each has five references. The resulting test set contains 3,805 captions. This differs from the standard 1,000-image Flickr30k gallery.

\paragraph{TextCaps.}
We construct an image--text retrieval task from TextCaps~\citep{sidorov2020textcaps}. Disjoint hash-selected pools of 6,000 training and 1,000 validation images come from the official training split. The official validation split supplies the held-out retrieval gallery of 3,166 images and 15,830 captions. We use CLIP image and caption encoders without an added OCR system. This experimental retrieval protocol differs from official caption evaluation.

\paragraph{Sampling, preprocessing, and metrics.}
For each seed, the 1,024-image adaptation set is a prefix of the 4,096-image set. Training batches sample distinct images and one reference caption per image. Retrieval uses deterministic CLIP resizing, center cropping, and normalization. Prompt positions precede the caption; words are retained up to CLIP's context limit. Exact decoded-image duplicates are removed across training and held-out partitions before model fitting. Identical splits and galleries are used by every method.

Evaluation ranks the complete gallery. Image-to-text success at rank $K$ requires any associated reference among the first $K$ captions; text-to-image success requires the annotated image among the first $K$ images. We report R@1, R@5, and R@10 in both directions. The primary measure averages the two R@1 values. Dataset-level averages weight datasets and training sizes equally rather than pooling queries across galleries.

\subsection{Language Diagnostic Scoring}
\label{app:language}
All diagnostic evaluations reuse terminal retrieval-trained models without updates. Each method contributes 30 models: five training datasets, two training-image counts, and three seeds. Retained cases are identical across methods and adaptation settings. We report the five benchmarks in Table~\ref{tab:language}, including all retained categories within each.

\paragraph{SugarCrepe.}
We use 7,511 pairs across all seven categories~\citep{hsieh2023sugarcrepe}: object/attribute addition, object/attribute/relation replacement, and object/attribute swapping. A pair is correct when $s(x,y^+)>s(x,y^-)$, with ties counted as failures. The score is the equal mean of the seven category accuracies.

\paragraph{CLEVR foils.}
From 2,000 official CLEVR validation images and scene annotations~\citep{CLEVR}, we construct 5,989 scene-verified pairs: 2,000 counting, 2,000 color-binding, and 1,989 relation-role-swap cases. Valid captions agree with the scene graph and foils violate the targeted property; ambiguous relations are excluded. We average the three category accuracies under strict positive-over-negative scoring. These diagnostics differ from CLEVR VQA and object-count classification.

\paragraph{SugarCrepe++.}
Each of the 4,757 retained cases has an image, two positive captions, and one negative~\citep{dumpala2024sugarcrepepp}. A case succeeds only when both positives strictly outrank the negative. We average accuracy across all five released categories: replacing attributes, objects, or relations, and swapping attributes or objects. SugarCrepe and SugarCrepe++ share source images, so their results are not independent samples.

\paragraph{What'sUp.}
We retain 4,713 of 4,958 cases from all six subsets~\citep{kamath2023whatsup}: Controlled A/B, COCO-QA one/two object, and VG-QA one/two object. Evaluation uses full images and the supplied caption choices. The official argmax rule places the correct caption first, so tied maxima favor it. The primary score equally averages individual accuracy across the six subsets. Controlled pair/set scores are secondary endpoints and do not replace individual accuracy in the main average.

\paragraph{EqBen.}
We evaluate all 25,104 image-pair cases in the author-released 10\% subset~\citep{wang2023eqben}; none is removed by the exact-image overlap filter. For paired images $(x_0,x_1)$ and captions $(y_0,y_1)$, write $s_{ij}=s(x_i,y_j)$. Text correctness requires $s_{00}>s_{01}$ and $s_{11}>s_{10}$; image correctness requires $s_{00}>s_{10}$ and $s_{11}>s_{01}$. The primary group score requires all four comparisons. Every tie fails. This group endpoint is stricter than single-caption choice and has a different chance level.

\paragraph{Overlap and averaging.}
For the newly downloaded diagnostics, we exclude the union of actual adaptation and validation images across all five training datasets. We use linked COCO/Flickr/Visual Genome identifiers when available and exact decoded-RGB hashes. A What'sUp controlled object-pair group is removed as a whole if any of its four conditions overlaps; an EqBen case is removed if either image overlaps. This audit does not test CLIP pretraining contamination or all perceptual duplicates. Earlier SugarCrepe and CLEVR exclusion checks were incorporated into the retrieval splits.

For each benchmark, scores are averaged over both training sizes and all five training datasets within each seed, then summarized by the mean and sample SD over seeds. The five-benchmark Average is computed within seed before calculating its SD. It is a descriptive equal-benchmark mean of different endpoint types, not a pooled accuracy. Diagnostic scores never select a model checkpoint or training hyperparameter.

\section{Implementation and Model Selection}
\label{app:protocol}
\subsection{Common Settings}
The new studies use CLIP ViT-B/16 with batch size 128 or ViT-L/14 with batch size 32. Unless ablated, ZOMP uses $T=4$ tokens in the first $L=9$ layers of both encoders and a two-stage schedule with $r_{\rm init}=1$, $r=4$, and $\rho=0.2$: rank one remains active through $q/Q=0.2$, with rank four used thereafter. We average up to $N=5$ two-sided probes per update and use SPSA-GC with $\beta=0.8$ and no gradient clipping. Gains are $c_k=c/k^{0.1}$ and $\eta_k=a/(1+k)^{0.4}$, corresponding to $\gamma=0.1$ and $\alpha=0.4$. Seeds 1, 2, and 3 control sampling and training randomness. Main zeroth-order runs have a 5,000-query cap, regardless of whether they use 1,024 or 4,096 training images.

We initialize shared factors with $\sigma_U=0.05$ and modality-specific factors to zero. Active perturbation entries are uniform on $[-1,-\tfrac12]\cup[\tfrac12,1]$. For ViT-B/16, $d_v=768$ and $d_t=512$, so Equation~\ref{eq:dimension} gives 11,556 active coordinates at rank one and 46,224 at rank four. Direct prompts at the same depth and token count contain 46,080 coordinates. The default reduces the early search dimension while retaining the expressivity of direct prompts at final rank; it does not reduce the final parameter count.

Rank components are defined by the initialized factors, without a singular-value decomposition or an assumption that the first component is a dominant singular direction. Activation preserves prompt tensors at the transition, but losses on different minibatches need not be continuous.

The controlled classification suite uses existing 16-shot training subsets and at most four validation examples per class. DTD supplies classification development, and the selected gains transfer to the other three tasks. Retrieval gains are developed per dataset and then shared across both training sizes and all main seeds. Development uses seed 1, 500 loss queries for zeroth-order methods, and five epochs for AdamW. Retrieval uses 4,096 development images.

\subsection{Development Grids}
The factor/direct/ZIP grid is $a\in\{0.003,0.01,0.03\}$ crossed with $c\in\{0.005,0.02\}$. Factor-family choices in the initial controlled study maximize the paired validation score of scheduled and fixed-rank configurations. The later Flickr30k/TextCaps extension develops ZOMP's factor gains directly. AdamW searches learning rates 0.00003, 0.0001, 0.0003, 0.001, 0.003, and 0.01. BlackVIP searches $(a,c)$ pairs $(0.003,0.01)$, $(0.01,0.005)$, $(0.01,0.01)$, $(0.01,0.02)$, $(0.03,0.01)$, and $(0.03,0.02)$. BAR crosses $a\in\{0.1,1,10\}$ with $c\in\{0.01,0.1\}$. Ties use the earliest grid index. BPT retains its native optimizer configuration.

\begin{table*}[!htbp]
\caption{Frozen development choices. A dash denotes an unrun configuration. Factor-family controls reuse their dataset's factor gains; plain-SPSA controls are not independently retuned.}
\label{tab:gains}
\centering\tablestyle
\begin{tabular*}{\linewidth}{@{\extracolsep{\fill}}lccccc@{}}
\toprule
\multicolumn{1}{c}{Task} & Factor $(a,c)$ & ZIP $(a,c)$ & BAR $(a,c)$ & Direct $(a,c)$ & AdamW LR\\
\midrule
DTD & $(0.01,0.005)$ & $(0.01,0.005)$ & -- & $(0.003,0.02)$ & $0.003$\\
COCO & $(0.003,0.005)$ & $(0.03,0.02)$ & $(10,0.1)$ & $(0.003,0.02)$ & $0.001$\\
Flickr8k & $(0.01,0.02)$ & $(0.01,0.005)$ & $(1,0.01)$ & -- & $0.01$\\
RSICD & $(0.003,0.02)$ & $(0.01,0.005)$ & $(1,0.01)$ & -- & $0.01$\\
Flickr30k & $(0.01,0.02)$ & $(0.01,0.005)$ & $(10,0.01)$ & -- & --\\
TextCaps & $(0.01,0.02)$ & $(0.01,0.02)$ & $(10,0.01)$ & -- & --\\
\bottomrule
\end{tabular*}
\end{table*}
BlackVIP uses $(a,c)=(0.003,0.01)$ on all five retrieval datasets. The combined development campaigns contain 204 runs: 144 in the initial studies and 60 for Flickr30k/TextCaps. Development costs are additional to each main run's budget. Validation selects hyperparameters; every primary table evaluates the terminal training checkpoint, including methods whose earlier validation score was higher.

\subsection{Baseline Implementations}
\paragraph{ZIP.}
We retain eight text tokens, intrinsic dimension 500, rank five, frozen Fastfood projection buffers, and native clipped SPSA. The caption adapter reserves positions before the sentence, and the visual branch remains unprompted. Retrieval replaces the classification objective with Equation~\ref{eq:retrieval}.

\paragraph{BPT-VLM.}
The retrieval comparison uses eight text tokens, five visual tokens, all 12 prompt layers for ViT-B/16, and layerwise full-covariance CMA-ES with 500 intrinsic coordinates per branch per layer. Backbone loading honors the common comparison backbone. Hidden-state calibration uses training images only. Candidate fitness averages loss over the adaptation set; every constituent minibatch is charged. An incomplete candidate is not evaluated. The terminal model is the final training incumbent after the last complete candidate.

\paragraph{BlackVIP.}
A frozen MAE encoder and trainable coordinator/decoder generate an image-dependent visual input prompt with 9,920 trainable coordinates. We retain SPSA-GC. Batch-normalization statistics remain frozen across perturbation evaluations, preventing state changes between probes.

\paragraph{BAR.}
Retrieval uses its native one-sided randomized gradient-free update: one pivot loss and five Gaussian probes, each charged to the budget. A $194\times194$ image is placed inside a $224\times224$ learnable frame. Its native input preprocessing differs from the unprompted CLIP crop.

\paragraph{First-order and matched controls.}
AdamW trains the same deep shared-factor architecture, with all rank-four coordinates active, for 20 epochs. It is not query-matched to zeroth-order methods. Modality controls bypass prompt insertion in the disabled branch. Depth-one and depth-nine controls otherwise share the factor-family settings. Direct optimization learns the prompt matrices at the same positions with SPSA-GC. Plain-SPSA controls omit GC and clipping.

\section{Complete Retrieval Results}
\label{app:retrieval}
Tables~\ref{tab:full_coco}--\ref{tab:full_textcaps} give both retrieval directions at ranks 1, 5, and 10. Values are mean and sample SD over three terminal checkpoints. Bold and underline compare the five adapted derivative-free methods within each block; AdamW is a separate first-order reference where available. Every derivative-free block has a 5,000-query cap, while the block label gives the number of training images.
\begin{table*}[!htbp]
\caption{COCO retrieval (\%). $I\to T$ retrieves captions; $T\to I$ retrieves images.}
\label{tab:full_coco}
\centering\tablestyle
\begin{tabular*}{\linewidth}{@{\extracolsep{\fill}}lcccccc>{\columncolor[gray]{0.9}}c@{}}
\toprule
\multicolumn{1}{c}{Method} & I$\to$T R@1 & R@5 & R@10 & T$\to$I R@1 & R@5 & R@10 & Mean R@1 \\
\midrule
\multicolumn{8}{c}{\textit{1,024 training images; 5,000 loss queries for derivative-free methods}} \\
BAR & 46.85\std{0.49} & 71.67\std{0.30} & 80.75\std{0.23} & 33.08\std{0.19} & 57.55\std{0.19} & 68.26\std{0.19} & 39.96\std{0.20} \\
\textsc{BlackVIP} & 49.41\std{0.50} & 73.47\std{0.13} & 81.77\std{0.13} & 32.97\std{0.51} & 57.72\std{0.50} & 68.26\std{0.35} & 41.19\std{0.43} \\
BPT-VLM & \underline{54.56}\std{0.40} & \underline{78.05}\std{0.57} & \underline{85.84}\std{0.39} & \underline{33.56}\std{0.39} & \underline{58.59}\std{0.21} & \underline{69.40}\std{0.08} & \underline{44.06}\std{0.19} \\
\textsc{ZIP} & 52.04\std{0.49} & 76.33\std{0.33} & 84.13\std{0.49} & 32.77\std{1.59} & 57.59\std{1.17} & 68.07\std{1.27} & 42.41\std{0.84} \\
\textbf{ZOMP} & \textbf{56.51}\std{0.11} & \textbf{79.47}\std{0.34} & \textbf{86.96}\std{0.42} & \textbf{36.55}\std{0.30} & \textbf{61.69}\std{0.28} & \textbf{72.16}\std{0.19} & \textbf{46.53}\std{0.10} \\
AdamW (20 epochs) & 58.10\std{0.61} & 80.83\std{0.64} & 88.19\std{0.29} & 39.97\std{0.72} & 65.99\std{0.79} & 76.05\std{0.64} & 49.04\std{0.62} \\
\midrule
\multicolumn{8}{c}{\textit{4,096 training images; 5,000 loss queries for derivative-free methods}} \\
BAR & 47.49\std{0.05} & 72.25\std{0.13} & 81.11\std{0.18} & 32.96\std{0.05} & 57.47\std{0.08} & 68.07\std{0.02} & 40.23\std{0.04} \\
\textsc{BlackVIP} & 49.73\std{0.28} & 73.64\std{0.42} & 82.21\std{0.09} & 32.90\std{0.31} & 57.69\std{0.30} & 68.31\std{0.30} & 41.32\std{0.14} \\
BPT-VLM & 54.19\std{0.48} & 77.27\std{0.29} & 84.95\std{0.08} & 33.33\std{0.02} & 57.95\std{0.19} & 68.52\std{0.27} & 43.76\std{0.25} \\
\textsc{ZIP} & \underline{54.21}\std{1.52} & \underline{77.89}\std{1.46} & \underline{86.10}\std{1.06} & \underline{33.58}\std{0.59} & \underline{58.33}\std{0.55} & \underline{68.91}\std{0.72} & \underline{43.90}\std{0.71} \\
\textbf{ZOMP} & \textbf{56.42}\std{0.23} & \textbf{79.19}\std{0.34} & \textbf{86.88}\std{0.35} & \textbf{36.31}\std{0.25} & \textbf{61.46}\std{0.41} & \textbf{71.96}\std{0.23} & \textbf{46.37}\std{0.19} \\
AdamW (20 epochs) & 60.00\std{0.54} & 82.43\std{0.51} & 89.35\std{0.09} & 43.08\std{0.46} & 69.23\std{0.47} & 78.91\std{0.42} & 51.54\std{0.48} \\
\bottomrule
\end{tabular*}

\end{table*}
\begin{table*}[!htbp]
\caption{Flickr8k retrieval (\%) under the shared modified validation protocol.}
\label{tab:full_flickr}
\centering\tablestyle
\begin{tabular*}{\linewidth}{@{\extracolsep{\fill}}lcccccc>{\columncolor[gray]{0.9}}c@{}}
\toprule
\multicolumn{1}{c}{Method} & I$\to$T R@1 & R@5 & R@10 & T$\to$I R@1 & R@5 & R@10 & Mean R@1 \\
\midrule
\multicolumn{8}{c}{\textit{1,024 training images; 5,000 loss queries for derivative-free methods}} \\
BAR & 69.70\std{0.36} & 89.87\std{0.29} & 95.43\std{0.06} & 55.59\std{0.34} & 82.61\std{0.19} & 90.15\std{0.06} & 62.64\std{0.32} \\
\textsc{BlackVIP} & 70.30\std{0.52} & 91.47\std{0.21} & 96.07\std{0.23} & 55.69\std{0.28} & 82.41\std{0.16} & 90.02\std{0.39} & 62.99\std{0.40} \\
BPT-VLM & \underline{73.77}\std{1.30} & 92.77\std{0.21} & 96.90\std{0.53} & \underline{56.77}\std{0.25} & \underline{83.51}\std{0.36} & \underline{91.44}\std{0.25} & \underline{65.27}\std{0.73} \\
\textsc{ZIP} & 73.43\std{1.34} & \underline{93.03}\std{1.27} & \underline{96.93}\std{0.35} & 55.40\std{0.62} & 82.21\std{0.78} & 89.92\std{0.50} & 64.42\std{0.86} \\
\textbf{ZOMP} & \textbf{76.57}\std{0.32} & \textbf{94.20}\std{1.14} & \textbf{97.40}\std{0.35} & \textbf{60.78}\std{0.49} & \textbf{85.77}\std{0.34} & \textbf{93.06}\std{0.25} & \textbf{68.67}\std{0.26} \\
AdamW (20 epochs) & 81.23\std{1.12} & 95.70\std{0.36} & 98.03\std{0.29} & 66.41\std{0.36} & 89.63\std{0.17} & 95.00\std{0.16} & 73.82\std{0.74} \\
\midrule
\multicolumn{8}{c}{\textit{4,096 training images; 5,000 loss queries for derivative-free methods}} \\
BAR & 68.97\std{0.06} & 89.83\std{0.25} & 95.33\std{0.21} & 55.27\std{0.19} & 82.48\std{0.17} & 89.94\std{0.06} & 62.12\std{0.11} \\
\textsc{BlackVIP} & 70.07\std{0.06} & 91.47\std{0.35} & 95.90\std{0.62} & 55.18\std{0.57} & 82.04\std{0.24} & 89.80\std{0.22} & 62.62\std{0.31} \\
BPT-VLM & \underline{74.50}\std{0.56} & 92.30\std{0.30} & 96.90\std{0.61} & \underline{56.25}\std{0.28} & \underline{83.01}\std{0.10} & \underline{90.83}\std{0.49} & \underline{65.37}\std{0.36} \\
\textsc{ZIP} & 73.33\std{0.70} & \underline{93.00}\std{0.30} & \underline{97.13}\std{0.49} & 55.30\std{0.90} & 81.82\std{0.67} & 89.87\std{0.53} & 64.32\std{0.72} \\
\textbf{ZOMP} & \textbf{76.33}\std{0.91} & \textbf{94.03}\std{0.57} & \textbf{97.53}\std{0.06} & \textbf{61.14}\std{0.73} & \textbf{86.15}\std{0.28} & \textbf{93.19}\std{0.16} & \textbf{68.74}\std{0.66} \\
AdamW (20 epochs) & 83.17\std{1.14} & 96.47\std{0.31} & 98.40\std{0.10} & 69.09\std{0.34} & 91.06\std{0.09} & 96.11\std{0.20} & 76.13\std{0.72} \\
\bottomrule
\end{tabular*}

\end{table*}
\begin{table*}[!htbp]
\caption{RSICD retrieval (\%). Relevance follows the supplied image--caption associations.}
\label{tab:full_rsicd}
\centering\tablestyle
\begin{tabular*}{\linewidth}{@{\extracolsep{\fill}}lcccccc>{\columncolor[gray]{0.9}}c@{}}
\toprule
\multicolumn{1}{c}{Method} & I$\to$T R@1 & R@5 & R@10 & T$\to$I R@1 & R@5 & R@10 & Mean R@1 \\
\midrule
\multicolumn{8}{c}{\textit{1,024 training images; 5,000 loss queries for derivative-free methods}} \\
BAR & 4.18\std{0.28} & 12.96\std{0.29} & 21.59\std{0.18} & 5.06\std{0.10} & \underline{17.27}\std{0.12} & \underline{28.48}\std{0.28} & 4.62\std{0.13} \\
\textsc{BlackVIP} & 4.51\std{0.14} & 13.39\std{0.46} & 22.54\std{0.78} & 4.65\std{0.37} & 16.19\std{0.53} & 26.95\std{0.47} & 4.58\std{0.17} \\
BPT-VLM & 6.19\std{0.88} & 17.32\std{1.32} & 27.26\std{0.66} & \underline{5.15}\std{0.28} & 17.18\std{0.72} & 27.98\std{0.33} & 5.67\std{0.57} \\
\textsc{ZIP} & \underline{6.92}\std{0.43} & \underline{18.60}\std{0.60} & \underline{27.51}\std{1.22} & 4.86\std{0.26} & 16.63\std{0.87} & 27.61\std{0.99} & \underline{5.89}\std{0.24} \\
\textbf{ZOMP} & \textbf{8.08}\std{0.85} & \textbf{21.04}\std{0.37} & \textbf{31.96}\std{0.85} & \textbf{5.76}\std{0.79} & \textbf{19.54}\std{0.53} & \textbf{32.08}\std{0.77} & \textbf{6.92}\std{0.52} \\
AdamW (20 epochs) & 10.00\std{0.45} & 24.21\std{1.27} & 35.65\std{1.09} & 8.31\std{0.52} & 26.69\std{0.67} & 41.74\std{1.20} & 9.16\std{0.27} \\
\midrule
\multicolumn{8}{c}{\textit{4,096 training images; 5,000 loss queries for derivative-free methods}} \\
BAR & 4.51\std{0.19} & 12.81\std{0.46} & 20.25\std{0.32} & 4.86\std{0.09} & 17.57\std{0.32} & 28.45\std{0.25} & 4.68\std{0.06} \\
\textsc{BlackVIP} & 5.06\std{0.41} & 14.15\std{0.37} & 23.54\std{0.28} & 4.99\std{0.17} & 16.87\std{0.65} & 27.67\std{0.98} & 5.03\std{0.28} \\
BPT-VLM & 5.70\std{0.37} & 16.74\std{0.88} & 26.04\std{0.65} & \underline{5.66}\std{0.08} & \underline{17.95}\std{0.34} & \underline{28.79}\std{0.23} & 5.68\std{0.22} \\
\textsc{ZIP} & \underline{6.71}\std{0.83} & \underline{17.78}\std{0.50} & \underline{27.60}\std{1.06} & 5.06\std{0.03} & 17.19\std{0.45} & 28.07\std{1.12} & \underline{5.89}\std{0.41} \\
\textbf{ZOMP} & \textbf{8.17}\std{0.14} & \textbf{20.89}\std{0.86} & \textbf{32.24}\std{0.61} & \textbf{5.70}\std{0.34} & \textbf{19.10}\std{0.50} & \textbf{31.78}\std{0.64} & \textbf{6.94}\std{0.13} \\
AdamW (20 epochs) & 13.78\std{0.19} & 31.11\std{1.28} & 43.28\std{1.76} & 10.78\std{0.40} & 31.48\std{0.13} & 47.37\std{1.07} & 12.28\std{0.14} \\
\bottomrule
\end{tabular*}

\end{table*}
\begin{table*}[!htbp]
\caption{Flickr30k retrieval (\%) on the filtered 761-image test gallery.}
\label{tab:full_flickr30k}
\centering\tablestyle
\begin{tabular*}{\linewidth}{@{\extracolsep{\fill}}lcccccc>{\columncolor[gray]{0.9}}c@{}}
\toprule
\multicolumn{1}{c}{Method} & I$\to$T R@1 & R@5 & R@10 & T$\to$I R@1 & R@5 & R@10 & Mean R@1 \\
\midrule
\multicolumn{8}{c}{\textit{1,024 training images; 5,000 loss queries for derivative-free methods}} \\
BAR & 82.48\std{0.42} & 96.01\std{0.33} & 98.60\std{0.08} & 62.80\std{0.12} & 86.40\std{0.14} & 91.92\std{0.07} & 72.64\std{0.26} \\
\textsc{BlackVIP} & 83.27\std{1.13} & 96.45\std{0.39} & 98.03\std{0.35} & 64.28\std{0.19} & 86.25\std{0.13} & 91.87\std{0.18} & 73.78\std{0.58} \\
BPT-VLM & 84.36\std{0.73} & \underline{97.42}\std{0.38} & \underline{99.12}\std{0.42} & \underline{65.58}\std{0.34} & \underline{87.38}\std{0.77} & \underline{92.55}\std{0.50} & \underline{74.97}\std{0.20} \\
\textsc{ZIP} & \underline{84.45}\std{0.46} & 96.58\std{0.35} & 98.64\std{0.27} & 64.71\std{0.58} & 86.61\std{0.14} & 91.76\std{0.25} & 74.58\std{0.13} \\
\textbf{ZOMP} & \textbf{86.99}\std{0.57} & \textbf{97.59}\std{0.08} & \textbf{99.21}\std{0.00} & \textbf{69.08}\std{0.32} & \textbf{89.91}\std{0.37} & \textbf{94.10}\std{0.39} & \textbf{78.03}\std{0.44} \\
\midrule
\multicolumn{8}{c}{\textit{4,096 training images; 5,000 loss queries for derivative-free methods}} \\
BAR & 82.57\std{0.40} & 95.97\std{0.27} & 98.60\std{0.20} & 62.79\std{0.30} & 86.12\std{0.09} & 92.00\std{0.22} & 72.68\std{0.20} \\
\textsc{BlackVIP} & 82.70\std{0.87} & 96.36\std{0.53} & 97.94\std{0.27} & 64.28\std{0.40} & 86.21\std{0.21} & 91.67\std{0.26} & 73.49\std{0.52} \\
BPT-VLM & 84.32\std{1.45} & 97.11\std{0.57} & 98.82\std{0.23} & \underline{65.60}\std{0.30} & \underline{87.14}\std{0.24} & \underline{92.15}\std{0.16} & \underline{74.96}\std{0.85} \\
\textsc{ZIP} & \underline{85.06}\std{1.97} & \underline{97.33}\std{0.50} & \underline{98.95}\std{0.23} & 64.50\std{0.68} & 86.32\std{0.33} & 91.48\std{0.20} & 74.78\std{1.30} \\
\textbf{ZOMP} & \textbf{88.22}\std{0.99} & \textbf{97.50}\std{0.13} & \textbf{99.12}\std{0.20} & \textbf{69.08}\std{0.28} & \textbf{90.05}\std{0.27} & \textbf{93.95}\std{0.17} & \textbf{78.65}\std{0.50} \\
\bottomrule
\end{tabular*}

\end{table*}
\begin{table*}[!htbp]
\caption{TextCaps retrieval (\%) on the 3,166-image held-out gallery.}
\label{tab:full_textcaps}
\centering\tablestyle
\begin{tabular*}{\linewidth}{@{\extracolsep{\fill}}lcccccc>{\columncolor[gray]{0.9}}c@{}}
\toprule
\multicolumn{1}{c}{Method} & I$\to$T R@1 & R@5 & R@10 & T$\to$I R@1 & R@5 & R@10 & Mean R@1 \\
\midrule
\multicolumn{8}{c}{\textit{1,024 training images; 5,000 loss queries for derivative-free methods}} \\
BAR & 70.59\std{0.27} & 85.88\std{0.18} & 90.81\std{0.05} & 53.89\std{0.11} & 71.64\std{0.07} & 78.47\std{0.01} & 62.24\std{0.17} \\
\textsc{BlackVIP} & 70.95\std{0.31} & 85.85\std{0.38} & 90.79\std{0.25} & 52.96\std{0.37} & 71.12\std{0.29} & 78.01\std{0.41} & 61.96\std{0.34} \\
BPT-VLM & \underline{76.57}\std{0.26} & \underline{89.02}\std{0.41} & \underline{93.02}\std{0.11} & \underline{56.69}\std{0.95} & \underline{74.39}\std{0.78} & \underline{80.63}\std{0.84} & \underline{66.63}\std{0.52} \\
\textsc{ZIP} & 75.69\std{0.87} & 88.86\std{0.41} & 92.71\std{0.29} & 55.96\std{0.58} & 73.59\std{0.40} & 79.92\std{0.30} & 65.82\std{0.52} \\
\textbf{ZOMP} & \textbf{78.23}\std{0.51} & \textbf{90.57}\std{0.10} & \textbf{94.07}\std{0.10} & \textbf{59.65}\std{0.27} & \textbf{77.09}\std{0.17} & \textbf{83.41}\std{0.21} & \textbf{68.94}\std{0.12} \\
\midrule
\multicolumn{8}{c}{\textit{4,096 training images; 5,000 loss queries for derivative-free methods}} \\
BAR & 70.74\std{0.16} & 85.90\std{0.07} & 90.68\std{0.17} & 53.88\std{0.06} & 71.69\std{0.14} & 78.37\std{0.04} & 62.31\std{0.06} \\
\textsc{BlackVIP} & 71.26\std{0.60} & 86.67\std{0.16} & 91.25\std{0.03} & 53.44\std{0.12} & 71.49\std{0.14} & 78.14\std{0.17} & 62.35\std{0.36} \\
BPT-VLM & \underline{75.96}\std{0.25} & 88.77\std{0.27} & \underline{92.63}\std{0.14} & \underline{56.72}\std{0.16} & \underline{74.69}\std{0.01} & \underline{81.02}\std{0.32} & \underline{66.34}\std{0.09} \\
\textsc{ZIP} & 75.24\std{0.52} & \underline{88.78}\std{0.56} & 92.48\std{0.28} & 56.07\std{0.81} & 73.75\std{0.78} & 80.15\std{0.85} & 65.66\std{0.65} \\
\textbf{ZOMP} & \textbf{78.58}\std{0.76} & \textbf{90.27}\std{0.36} & \textbf{93.98}\std{0.23} & \textbf{59.81}\std{0.38} & \textbf{77.24}\std{0.56} & \textbf{83.48}\std{0.39} & \textbf{69.20}\std{0.28} \\
\bottomrule
\end{tabular*}

\end{table*}

\subsection{Paired Uncertainty Estimates}
\label{app:intervals}
We use 5,000 paired bootstrap replicates. Image-cluster resampling retains associated captions, and hierarchical intervals resample the three training seeds before the image clusters. Table~\ref{tab:intervals} gives the initial six retrieval settings referenced in the main text. These descriptive intervals have limited seed coverage and no multiplicity adjustment.
\begin{table*}[!htbp]
\caption{ZOMP minus each external comparator in mean bidirectional R@1 (points), with paired seed/image-cluster 95\% intervals. All runs use terminal checkpoints under a 5,000-query cap.}
\label{tab:intervals}
\centering\tablestyle
\begin{tabular*}{\linewidth}{@{\extracolsep{\fill}}lcccc@{}}
\toprule
\multicolumn{1}{c}{Training dataset / images} & BAR & BlackVIP & ZIP & BPT-VLM \\
\midrule
COCO / 1,024 & +6.57 [+5.85, +7.27] & +5.34 [+4.60, +6.08] & +4.12 [+3.27, +5.05] & +2.47 [+1.94, +3.03] \\
COCO / 4,096 & +6.14 [+5.41, +6.85] & +5.05 [+4.30, +5.80] & +2.47 [+1.47, +3.34] & +2.60 [+2.11, +3.09] \\
Flickr8k / 1,024 & +6.03 [+4.49, +7.59] & +5.68 [+4.19, +7.15] & +4.26 [+2.99, +5.56] & +3.41 [+2.18, +4.59] \\
Flickr8k / 4,096 & +6.62 [+5.04, +8.20] & +6.11 [+4.57, +7.65] & +4.42 [+3.29, +5.54] & +3.36 [+2.02, +4.72] \\
RSICD / 1,024 & +2.30 [+1.13, +3.43] & +2.34 [+1.37, +3.35] & +1.03 [-0.07, +2.15] & +1.25 [+0.06, +2.49] \\
RSICD / 4,096 & +2.25 [+1.25, +3.25] & +1.91 [+0.94, +2.86] & +1.05 [+0.15, +1.99] & +1.25 [+0.31, +2.16] \\
\bottomrule
\end{tabular*}

\end{table*}

\section{Diagnostics by Adaptation Dataset}
\label{app:language_results}
\begin{figure*}[!htbp]
\centering\includegraphics[width=\textwidth]{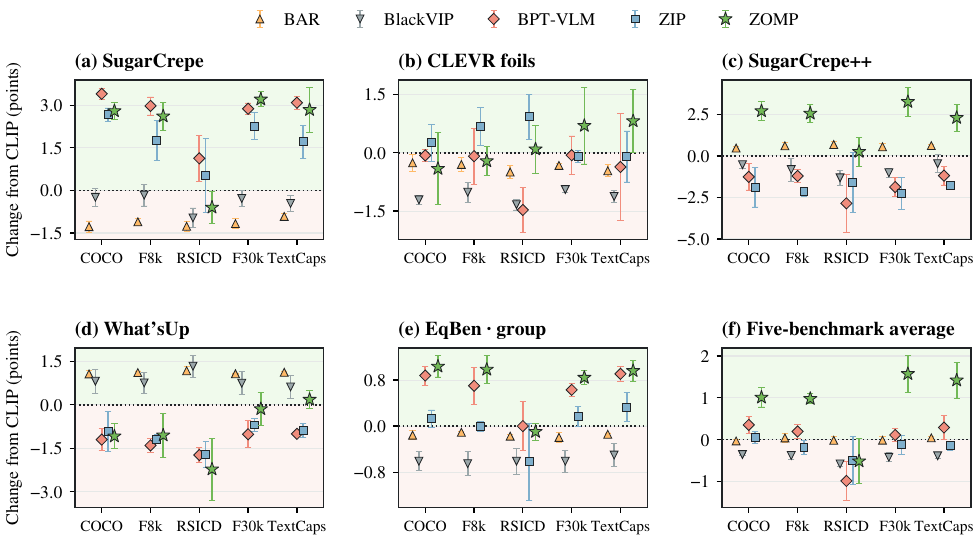}
\caption{\textbf{Language transfer across adaptation domains.} Change from unprompted CLIP on the five selected diagnostics and their equal-benchmark average. Each point averages both training-image counts; error bars give sample SD over three seeds. F8k and F30k denote Flickr8k and Flickr30k. Green and coral regions indicate positive and negative changes. All five adaptation datasets and four external baselines are included; the benchmark subset is the exploratory selection in Table~\ref{tab:language}.}
\label{fig:language_transfer}
\end{figure*}

Tables~\ref{tab:lang_groups1} and~\ref{tab:lang_groups2} separate the training-dataset groups underlying Table~\ref{tab:language}. Each entry averages both training-image counts within seed, then reports mean and sample SD across seeds. Benchmark definitions and retained cases are unchanged across groups. These models were trained for retrieval, not on the diagnostic labels. Figure~\ref{fig:language_transfer} shows changes from unprompted CLIP by adaptation domain.
\begin{table*}[!htbp]
\caption{Selected five-benchmark diagnostics after COCO, Flickr8k, or RSICD adaptation. Benchmark columns use the metrics in Appendix~\ref{app:language}; EqBen is the official 10\% subset's group score. All runs use a 5,000-query cap.}
\label{tab:lang_groups1}
\centering\tablestyle
\begin{tabular*}{\linewidth}{@{\extracolsep{\fill}}lccccc>{\columncolor[gray]{0.9}}c@{}}
\toprule
\multicolumn{1}{c}{Method} & SugarCrepe & CLEVR foils & SugarCrepe++ & What'sUp & EqBen$^{\dagger}$ & Average \\
\midrule
\multicolumn{7}{c}{\textit{COCO adaptation}} \\
BAR & 71.82\std{0.20} & 58.67\std{0.22} & \underline{60.19}\std{0.12} & \textbf{44.01}\std{0.10} & 10.13\std{0.08} & 48.96\std{0.03} \\
\textsc{BlackVIP} & 72.85\std{0.32} & 57.72\std{0.12} & 59.18\std{0.22} & \underline{43.76}\std{0.41} & 9.68\std{0.17} & 48.64\std{0.02} \\
BPT-VLM & \textbf{76.50}\std{0.19} & \underline{58.87}\std{0.15} & 58.45\std{0.79} & 41.75\std{0.38} & \underline{11.16}\std{0.16} & \underline{49.35}\std{0.20} \\
\textsc{ZIP} & 75.76\std{0.24} & \textbf{59.19}\std{0.48} & 57.82\std{1.18} & 42.02\std{0.68} & 10.42\std{0.15} & 49.04\std{0.14} \\
\textbf{ZOMP} & \underline{75.89}\std{0.31} & 58.52\std{0.93} & \textbf{62.42}\std{0.56} & 41.86\std{0.42} & \textbf{11.32}\std{0.18} & \textbf{50.00}\std{0.23} \\
\midrule
\multicolumn{7}{c}{\textit{Flickr8k adaptation}} \\
BAR & 71.99\std{0.13} & 58.63\std{0.18} & \underline{60.32}\std{0.18} & \textbf{44.06}\std{0.06} & 10.18\std{0.02} & 49.04\std{0.10} \\
\textsc{BlackVIP} & 72.93\std{0.38} & 57.92\std{0.25} & 58.89\std{0.69} & \underline{43.70}\std{0.36} & 9.63\std{0.21} & 48.62\std{0.10} \\
BPT-VLM & \textbf{76.07}\std{0.32} & \underline{58.84}\std{0.72} & 58.51\std{0.39} & 41.54\std{0.25} & \underline{10.99}\std{0.32} & \underline{49.19}\std{0.16} \\
\textsc{ZIP} & 74.86\std{0.70} & \textbf{59.61}\std{0.49} & 57.55\std{0.28} & 41.76\std{0.15} & 10.28\std{0.08} & 48.81\std{0.17} \\
\textbf{ZOMP} & \underline{75.71}\std{0.50} & 58.72\std{0.37} & \textbf{62.27}\std{0.54} & 41.90\std{0.75} & \textbf{11.27}\std{0.25} & \textbf{49.97}\std{0.12} \\
\midrule
\multicolumn{7}{c}{\textit{RSICD adaptation}} \\
BAR & 71.83\std{0.18} & 58.44\std{0.16} & \textbf{60.40}\std{0.19} & \underline{44.12}\std{0.14} & 10.11\std{0.04} & \textbf{48.98}\std{0.09} \\
\textsc{BlackVIP} & 72.13\std{0.35} & 57.60\std{0.14} & 58.40\std{0.44} & \textbf{44.28}\std{0.38} & 9.68\std{0.22} & 48.42\std{0.05} \\
BPT-VLM & \textbf{74.23}\std{0.81} & 57.46\std{0.58} & 56.86\std{1.74} & 41.21\std{0.26} & \textbf{10.29}\std{0.43} & 48.01\std{0.46} \\
\textsc{ZIP} & \underline{73.62}\std{1.30} & \textbf{59.86}\std{0.58} & 58.13\std{1.80} & 41.24\std{0.45} & 9.66\std{0.67} & \underline{48.50}\std{0.58} \\
\textbf{ZOMP} & 72.50\std{0.58} & \underline{59.03}\std{0.62} & \underline{59.97}\std{0.87} & 40.73\std{1.07} & \underline{10.19}\std{0.15} & 48.48\std{0.54} \\
\bottomrule
\end{tabular*}

\end{table*}
\begin{table*}[!htbp]
\caption{The same diagnostic endpoints after Flickr30k or TextCaps adaptation. Average gives equal weight to the five benchmark columns.}
\label{tab:lang_groups2}
\centering\tablestyle
\begin{tabular*}{\linewidth}{@{\extracolsep{\fill}}lccccc>{\columncolor[gray]{0.9}}c@{}}
\toprule
\multicolumn{1}{c}{Method} & SugarCrepe & CLEVR foils & SugarCrepe++ & What'sUp & EqBen$^{\dagger}$ & Average \\
\midrule
\multicolumn{7}{c}{\textit{Flickr30k$^{\dagger}$ adaptation}} \\
BAR & 71.93\std{0.18} & 58.61\std{0.04} & \underline{60.27}\std{0.21} & \textbf{44.02}\std{0.03} & 10.09\std{0.08} & 48.98\std{0.08} \\
\textsc{BlackVIP} & 72.81\std{0.28} & 57.99\std{0.08} & 58.70\std{0.28} & \underline{43.69}\std{0.40} & 9.68\std{0.19} & 48.57\std{0.11} \\
BPT-VLM & \underline{75.97}\std{0.20} & \underline{58.87}\std{0.49} & 57.84\std{0.60} & 41.93\std{0.47} & \underline{10.92}\std{0.12} & \underline{49.11}\std{0.16} \\
\textsc{ZIP} & 75.36\std{0.46} & 58.84\std{0.16} & 57.45\std{0.97} & 42.25\std{0.23} & 10.45\std{0.17} & 48.87\std{0.22} \\
\textbf{ZOMP} & \textbf{76.31}\std{0.26} & \textbf{59.63}\std{0.98} & \textbf{62.97}\std{0.86} & 42.80\std{0.57} & \textbf{11.13}\std{0.12} & \textbf{50.57}\std{0.44} \\
\midrule
\multicolumn{7}{c}{\textit{TextCaps$^{\dagger}$ adaptation}} \\
BAR & 72.18\std{0.09} & 58.47\std{0.16} & \underline{60.34}\std{0.04} & \textbf{44.07}\std{0.05} & 10.14\std{0.02} & 49.04\std{0.04} \\
\textsc{BlackVIP} & 72.64\std{0.27} & 57.81\std{0.15} & 59.25\std{0.54} & \underline{43.56}\std{0.40} & 9.78\std{0.20} & 48.61\std{0.06} \\
BPT-VLM & \textbf{76.18}\std{0.23} & 58.57\std{1.38} & 58.53\std{0.56} & 41.94\std{0.09} & \underline{11.19}\std{0.13} & \underline{49.28}\std{0.28} \\
\textsc{ZIP} & 74.81\std{0.58} & \underline{58.84}\std{0.66} & 57.93\std{0.18} & 42.05\std{0.24} & 10.62\std{0.25} & 48.85\std{0.10} \\
\textbf{ZOMP} & \underline{75.94}\std{0.79} & \textbf{59.75}\std{0.80} & \textbf{62.01}\std{0.81} & 43.13\std{0.32} & \textbf{11.25}\std{0.18} & \textbf{50.41}\std{0.43} \\
\bottomrule
\end{tabular*}

\end{table*}

\section{Controlled Classification and Backbone Results}
\label{app:controlled}
Table~\ref{tab:class_controls} compares five rank/optimizer configurations across all 13 classification datasets under the original native protocol: $(a,c)=(0.01,0.01)$, nine prompted layers, four tokens, and no gradient clipping. All configurations use 16 shots, a 5,000-query budget, and terminal checkpoints. The four rank-four configurations share the factor architecture; fixed rank one allocates rank-one factors throughout. Momentum denotes the SPSA-GC update; its removal retains the same gains and uses plain SPSA.

Each configuration uses seeds 1--3 on the first 12 datasets and seed 1 on ImageNet, for 185 runs. The same ImageNet seed is used for every configuration. Average equally weights the 13 dataset scores, with no SD reported for ImageNet or the mixed-seed overall mean. ZOMP reaches 68.24, compared with 67.32 for fixed rank one and 67.29 for fixed rank four with momentum. Its EuroSAT result is 80.45, consistent with Table~\ref{tab:fewshot}. Scheduling gains 0.95 points with momentum; removing momentum from ZOMP reduces the mean by 1.16 points. Under plain SPSA, scheduling changes 67.15 to 67.08. These comparisons assess component removal at fixed gains, rather than separately optimized algorithms.
\begin{table*}[!htbp]
\caption{\textbf{Rank and momentum ablation on all 13 datasets under the original protocol.} Test accuracy (\%) at 5,000 queries with $(a,c)=(0.01,0.01)$, terminal checkpoints, and no clipping. Entries are mean\,$\pm$\,SD over three seeds, except $\dagger$ImageNet uses seed 1 for every row. Average equally weights the 13 dataset scores; no SD is assigned to ImageNet or the mixed-seed average.}
\label{tab:class_controls}
\centering\densetablestyle
\begin{tabular*}{\linewidth}{@{\extracolsep{\fill}}l*{13}{c}>{\columncolor[gray]{0.9}}c@{}}
\toprule
\multicolumn{1}{c}{Method} & \rotatebox[origin=c]{90}{Caltech101} & \rotatebox[origin=c]{90}{OxfordPets} & \rotatebox[origin=c]{90}{Flowers102} & \rotatebox[origin=c]{90}{Food101} & \rotatebox[origin=c]{90}{FGVCAircraft} & \rotatebox[origin=c]{90}{SUN397} & \rotatebox[origin=c]{90}{DTD} & \rotatebox[origin=c]{90}{SVHN} & \rotatebox[origin=c]{90}{EuroSAT} & \rotatebox[origin=c]{90}{RESISC45} & \rotatebox[origin=c]{90}{CLEVR} & \rotatebox[origin=c]{90}{UCF101} & \rotatebox[origin=c]{90}{ImageNet$^\dagger$} & \rotatebox[origin=c]{90}{\textbf{Average}} \\
\midrule
Fixed $r=1$ + momentum & 94.3\std{0.6} & 92.9\std{0.3} & 73.5\std{1.8} & 87.0\std{0.1} & 27.1\std{1.6} & 67.8\std{0.3} & 55.4\std{2.5} & \underline{51.0}\std{0.9} & 76.6\std{2.2} & \underline{72.5}\std{0.3} & \textbf{36.4}\std{0.6} & 72.1\std{1.9} & 68.4 & \underline{67.32} \\
Fixed $r=4$, no momentum & 94.4\std{0.3} & \underline{93.1}\std{0.2} & 74.4\std{0.9} & \textbf{87.2}\std{0.1} & \underline{28.4}\std{0.8} & \textbf{68.4}\std{0.2} & 54.9\std{1.5} & 50.5\std{1.7} & 76.6\std{0.9} & 71.0\std{1.1} & 31.4\std{1.8} & \underline{73.2}\std{0.6} & \textbf{69.6} & 67.15 \\
Scheduled, no momentum & 94.4\std{0.3} & 92.9\std{0.4} & 74.6\std{1.7} & \underline{87.1}\std{0.1} & \textbf{28.9}\std{0.5} & \underline{68.1}\std{0.2} & \underline{55.8}\std{1.1} & 47.7\std{2.7} & \underline{79.1}\std{3.4} & \underline{72.5}\std{0.6} & 28.6\std{3.1} & 73.0\std{1.1} & \underline{69.4} & 67.08 \\
Fixed $r=4$ + momentum & \underline{94.7}\std{0.2} & \textbf{93.3}\std{0.3} & \underline{75.4}\std{0.7} & 87.0\std{0.1} & 25.4\std{2.5} & 67.5\std{0.4} & \underline{55.8}\std{1.6} & 49.9\std{5.1} & 77.6\std{6.5} & 72.3\std{1.5} & \underline{35.1}\std{2.1} & 72.3\std{0.8} & 68.5 & 67.29 \\
\textbf{ZOMP} & \textbf{94.8}\std{0.3} & 93.0\std{0.2} & \textbf{76.5}\std{1.3} & \underline{87.1}\std{0.1} & 28.2\std{0.5} & \textbf{68.4}\std{0.7} & \textbf{56.4}\std{1.3} & \textbf{51.4}\std{6.1} & \textbf{80.4}\std{1.8} & \textbf{73.5}\std{0.5} & \underline{35.1}\std{2.3} & \textbf{73.6}\std{1.5} & 68.7 & \textbf{68.24} \\
\bottomrule
\end{tabular*}

\end{table*}
\begin{table*}[!htbp]
\caption{Matched modality/depth controls at 5,000 queries. Classification uses the four-dataset development protocol with $(a,c)=(0.01,0.005)$; Table~\ref{tab:fewshot} uses $(0.01,0.01)$. COCO uses its retrieval gains $(0.003,0.005)$, mean R@1, and 4,096 training images. Class.\ avg.\ excludes COCO. A disabled branch has no inserted prompt.}
\label{tab:modality}
\centering\tablestyle
\begin{tabular*}{\linewidth}{@{\extracolsep{\fill}}lccccc>{\columncolor[gray]{0.9}}c@{}}
\toprule
\multicolumn{1}{c}{Method} & DTD & EuroSAT & Oxford Pets & Aircraft & COCO R@1 & Class. avg. \\
\midrule
Text only, $L=1$ & 51.10\std{0.94} & 62.65\std{5.06} & 92.72\std{0.42} & \textbf{27.49}\std{0.54} & 43.73\std{0.09} & 58.49\std{1.38} \\
Text only, $L=9$ & \underline{56.62}\std{0.74} & 65.76\std{7.48} & \textbf{93.27}\std{0.34} & \underline{27.19}\std{0.68} & \textbf{46.66}\std{0.36} & \underline{60.71}\std{2.08} \\
Vision only, $L=1$ & 46.02\std{0.76} & 59.26\std{8.22} & 88.50\std{0.33} & 22.89\std{1.96} & 45.32\std{0.40} & 54.17\std{2.56} \\
Vision only, $L=9$ & 46.38\std{0.87} & \textbf{78.90}\std{5.39} & 88.71\std{0.16} & 26.21\std{0.36} & 42.82\std{0.02} & 60.05\std{1.47} \\
Both branches, $L=1$ & 49.11\std{0.43} & 70.70\std{2.77} & 91.79\std{0.59} & 25.02\std{0.69} & 45.50\std{0.85} & 59.16\std{1.00} \\
Both branches, $L=9$ & \textbf{57.78}\std{1.81} & \underline{73.84}\std{6.29} & \underline{92.87}\std{0.40} & 27.04\std{0.61} & \underline{46.37}\std{0.19} & \textbf{62.88}\std{1.22} \\
\bottomrule
\end{tabular*}

\end{table*}
Table~\ref{tab:modality} varies modality and depth with gains and training settings held fixed within each task; Figure~\ref{fig:modality} visualizes the four-dataset classification mean. Its depth-nine dual row is ZOMP under the development protocol, including 73.84 on EuroSAT; the 80.45 result belongs to the original protocol in Table~\ref{tab:fewshot}. The depth-nine dual model leads the development-suite mean, while COCO favors text-only prompting. In the initial three-dataset retrieval controls, scheduling changes the mean over fixed-rank GC by $+0.13$ and $+0.04$ points at 1,024 and 4,096 training images. These small effects limit claims about the schedule's contribution to retrieval.

\begin{table*}[!htbp]
\begin{minipage}[t]{0.48\textwidth}
\caption{ViT-L/14 results with batch size 32. DTD/Pets use 16 shots; COCO uses 4,096 images. The derivative-free cap is 5,000 queries; AdamW uses 20 epochs.}
\label{tab:large}
\centering\tablestyle
\begin{tabular*}{\linewidth}{@{\extracolsep{\fill}}lccc@{}}
\toprule
\multicolumn{1}{c}{Method} & DTD & Oxford Pets & COCO R@1 \\
\midrule
\textsc{ZIP} & \underline{54.08}\std{1.74} & \underline{94.03}\std{0.37} & \underline{43.46}\std{0.35} \\
\textbf{ZOMP} & \textbf{59.63}\std{0.27} & \textbf{94.58}\std{0.20} & \textbf{48.19}\std{0.19} \\
AdamW (20 epochs) & 73.82\std{0.56} & 94.89\std{0.74} & 53.49\std{1.63} \\
\bottomrule
\end{tabular*}

\end{minipage}\hfill
\begin{minipage}[t]{0.48\textwidth}
\caption{COCO training resources: ViT-B/16, batch size 128, 4,096 training images. Memory is mean peak allocated training memory over seeds. AdamW uses a separate first-order horizon.}
\label{tab:memory}
\centering\tablestyle
\begin{tabular*}{\linewidth}{@{\extracolsep{\fill}}lcc@{}}
\toprule
\multicolumn{1}{c}{Method} & Peak allocated GiB & Loss queries \\
\midrule
BAR & 1.85 & 5000 \\
BlackVIP & 2.72 & 5000 \\
ZIP & 1.26 & 5000 \\
BPT-VLM & 2.24 & 4992 \\
ZOMP & 1.31 & 5000 \\
AdamW (20 epochs) & 15.52 & 20 epochs \\
\bottomrule
\end{tabular*}

\end{minipage}
\end{table*}

\section{Broad Classification Details}
\label{app:classification}
We evaluate ImageNet~\citep{deng2009imagenet}, Caltech101~\citep{caltech101}, Oxford Pets~\citep{oxford_pets}, Flowers102~\citep{flowers102}, Food101~\citep{food101}, Aircraft~\citep{aircraft}, SUN397~\citep{sun397}, DTD~\citep{dtd}, SVHN~\citep{SVHN}, EuroSAT~\citep{eurosat}, RESISC45~\citep{Resisc45}, CLEVR count classification~\citep{CLEVR}, and UCF101~\citep{ucf101}. Existing few-shot splits and class templates are retained, with 16 examples per class, three seeds, and 5,000 loss queries.

Broad-suite ZOMP uses $(a,c)=(0.01,0.01)$, nine layers, four tokens, and the 20\% unlock, with random resized crops and flips. The controlled suite develops gains on DTD and uses separately frozen code and random streams. Broad-suite BPT uses dual-encoder prompts with 2,000 intrinsic coordinates per modality. The broad input-frame reproduction uses SPSA-GC, while new retrieval BAR uses its native one-sided update. These protocol differences explain why overlapping datasets can have different scores; controlled results do not replace original suite entries.

\begin{table*}[!htbp]
\caption{Base and new accuracies underlying the main-text HM table. All 13 datasets, 16 shots, 5,000 queries; mean\,$\pm$\,SD. Ranking is within each block.}
\label{tab:broad_base}\label{tab:broad_new}
\centering\densetablestyle
\begin{tabular*}{\linewidth}{@{\extracolsep{\fill}}lccccccccccccc>{\columncolor[gray]{0.9}}c@{}}
\toprule
\multicolumn{1}{c}{Method} & \rotatebox[origin=c]{90}{Caltech101} & \rotatebox[origin=c]{90}{OxfordPets} & \rotatebox[origin=c]{90}{Flowers102} & \rotatebox[origin=c]{90}{Food101} & \rotatebox[origin=c]{90}{FGVCAircraft} & \rotatebox[origin=c]{90}{SUN397} & \rotatebox[origin=c]{90}{DTD} & \rotatebox[origin=c]{90}{SVHN} & \rotatebox[origin=c]{90}{EuroSAT} & \rotatebox[origin=c]{90}{RESISC45} & \rotatebox[origin=c]{90}{CLEVR} & \rotatebox[origin=c]{90}{UCF101} & \rotatebox[origin=c]{90}{ImageNet} & \rotatebox[origin=c]{90}{Average} \\
\midrule
\multicolumn{15}{c}{\textit{Base classes}} \\
BAR & 95.4\std{0.1} & 89.9\std{0.1} & 70.8\std{0.2} & 88.2\std{0.1} & 24.4\std{0.2} & 67.8\std{0.1} & 48.3\std{0.7} & 20.6\std{0.6} & 72.6\std{0.3} & 65.0\std{0.1} & 29.8\std{0.4} & 68.1\std{0.3} & 69.5\std{0.0} & 62.3\std{0.1} \\
\textsc{BlackVIP} & 96.4\std{0.2} & 91.0\std{0.2} & 71.8\std{0.7} & 88.2\std{0.1} & 26.9\std{0.6} & 68.7\std{0.5} & 52.9\std{0.7} & 34.5\std{6.7} & 72.2\std{2.4} & 71.2\std{0.9} & \underline{56.1}\std{0.6} & 71.2\std{0.7} & 70.2\std{0.2} & 67.0\std{0.7} \\
BPT-VLM & 94.2\std{2.0} & 92.8\std{1.1} & 67.7\std{2.3} & 88.9\std{0.5} & 29.7\std{0.3} & 63.0\std{2.3} & 56.9\std{0.6} & 49.1\std{2.8} & 78.8\std{1.8} & 74.8\std{1.3} & 45.3\std{6.6} & 70.6\std{2.9} & 67.3\std{1.5} & 67.6\std{0.3} \\
\textsc{ZIP} & \underline{97.0}\std{0.4} & \underline{94.9}\std{0.5} & \underline{72.2}\std{1.2} & \underline{90.2}\std{0.2} & \underline{30.6}\std{0.6} & \underline{69.4}\std{1.2} & \underline{60.6}\std{2.2} & \underline{57.7}\std{6.5} & \underline{83.6}\std{2.7} & \underline{83.6}\std{0.8} & 55.4\std{8.6} & \underline{73.6}\std{1.2} & \underline{72.2}\std{0.8} & \underline{72.4}\std{1.0} \\
\textbf{ZOMP} & \textbf{97.6}\std{0.1} & \textbf{95.3}\std{0.1} & \textbf{83.9}\std{1.4} & \textbf{90.6}\std{0.2} & \textbf{32.2}\std{0.5} & \textbf{74.7}\std{0.5} & \textbf{71.2}\std{1.6} & \textbf{58.3}\std{0.2} & \textbf{88.8}\std{2.2} & \textbf{86.6}\std{1.0} & \textbf{59.4}\std{1.1} & \textbf{78.6}\std{2.0} & \textbf{74.2}\std{0.3} & \textbf{76.3}\std{0.3} \\
\midrule
\multicolumn{15}{c}{\textit{New classes}} \\
BAR & \underline{94.1}\std{0.1} & 95.7\std{0.2} & \textbf{76.9}\std{0.2} & 89.4\std{0.2} & 31.3\std{0.2} & 74.1\std{0.2} & \underline{54.0}\std{1.6} & 34.4\std{1.0} & \textbf{81.2}\std{0.2} & 59.4\std{0.2} & \underline{31.2}\std{0.5} & 75.3\std{0.2} & 65.3\std{0.2} & \underline{66.3}\std{0.1} \\
\textsc{BlackVIP} & 93.2\std{0.3} & 95.2\std{0.2} & \underline{75.7}\std{0.5} & 89.9\std{0.2} & 33.4\std{0.9} & \underline{74.2}\std{0.4} & \textbf{57.3}\std{0.4} & 38.1\std{7.4} & 37.2\std{2.3} & 62.0\std{1.8} & 30.6\std{3.3} & \textbf{76.6}\std{0.5} & \underline{66.7}\std{0.1} & 63.8\std{0.5} \\
BPT-VLM & 92.4\std{0.5} & 94.4\std{3.0} & 73.8\std{2.9} & 89.1\std{0.5} & \underline{33.5}\std{3.2} & 63.7\std{5.6} & 46.9\std{0.6} & 38.3\std{0.4} & \underline{70.8}\std{10.2} & 60.8\std{2.0} & 23.6\std{1.9} & 68.6\std{2.1} & 58.0\std{2.5} & 62.6\std{1.7} \\
\textsc{ZIP} & \textbf{94.4}\std{1.2} & \underline{97.0}\std{0.6} & 72.6\std{0.9} & \underline{90.3}\std{1.0} & 30.2\std{1.7} & 71.5\std{1.8} & 52.3\std{3.6} & \underline{43.7}\std{5.2} & 60.0\std{7.0} & \underline{63.2}\std{1.0} & 24.5\std{0.6} & 71.6\std{3.8} & 65.5\std{1.8} & 64.4\std{0.8} \\
\textbf{ZOMP} & 94.0\std{1.1} & \textbf{97.5}\std{0.3} & 75.1\std{0.8} & \textbf{91.7}\std{0.1} & \textbf{34.1}\std{0.6} & \textbf{77.7}\std{0.6} & 53.5\std{2.7} & \textbf{45.6}\std{5.1} & 60.4\std{11.4} & \textbf{69.4}\std{1.1} & \textbf{34.8}\std{3.9} & \underline{75.6}\std{2.3} & \textbf{69.6}\std{0.1} & \textbf{67.6}\std{0.5} \\
\bottomrule
\end{tabular*}

\end{table*}
\begin{table*}[!htbp]
\caption{Cross-dataset transfer of ImageNet-trained prompts, with no target adaptation. Source is excluded from the 12-target Average. All five methods use a 5,000-query source-training cap; bold and underline compare adapted methods.}
\label{tab:cdt}\label{tab:broad_cdt}
\centering\densetablestyle
\begin{tabular*}{\linewidth}{@{\extracolsep{\fill}}lccccccccccccc>{\columncolor[gray]{0.9}}c@{}}
\toprule
\multicolumn{1}{c}{Method} & \rotatebox[origin=c]{90}{Source} & \rotatebox[origin=c]{90}{Caltech101} & \rotatebox[origin=c]{90}{OxfordPets} & \rotatebox[origin=c]{90}{Flowers102} & \rotatebox[origin=c]{90}{Food101} & \rotatebox[origin=c]{90}{FGVCAircraft} & \rotatebox[origin=c]{90}{SUN397} & \rotatebox[origin=c]{90}{DTD} & \rotatebox[origin=c]{90}{SVHN} & \rotatebox[origin=c]{90}{EuroSAT} & \rotatebox[origin=c]{90}{RESISC45} & \rotatebox[origin=c]{90}{CLEVR} & \rotatebox[origin=c]{90}{UCF101} & \rotatebox[origin=c]{90}{\textbf{Average}} \\
\midrule
BAR & 63.5\std{0.2} & 91.8\std{0.1} & 87.8\std{0.2} & \textbf{70.0}\std{0.3} & 83.7\std{0.1} & \underline{21.6}\std{0.4} & 60.9\std{0.2} & 39.4\std{0.3} & 19.3\std{0.3} & \textbf{61.0}\std{0.9} & 50.3\std{0.2} & 15.9\std{0.4} & 63.9\std{0.3} & \underline{55.5}\std{0.1} \\
\textsc{BlackVIP} & 64.6\std{0.3} & \underline{92.7}\std{0.1} & \textbf{88.1}\std{0.5} & \underline{68.8}\std{0.2} & 82.9\std{0.7} & \textbf{23.3}\std{0.2} & \underline{61.6}\std{0.4} & \underline{42.1}\std{0.5} & 15.3\std{0.9} & 36.5\std{2.6} & \underline{55.3}\std{0.5} & \underline{16.2}\std{0.8} & \underline{66.2}\std{0.6} & 54.1\std{0.5} \\
BPT-VLM & 58.7\std{1.2} & 82.0\std{3.1} & 76.9\std{1.9} & 54.7\std{5.1} & 76.2\std{4.5} & 15.7\std{0.5} & 41.5\std{6.6} & 27.5\std{4.5} & 14.6\std{3.1} & 38.7\std{7.4} & 36.4\std{2.2} & 12.2\std{0.5} & 50.4\std{7.2} & 43.9\std{2.8} \\
\textsc{ZIP} & \underline{65.8}\std{1.5} & 92.3\std{1.4} & \underline{87.9}\std{2.0} & 64.1\std{2.7} & \underline{84.3}\std{1.3} & 19.3\std{1.4} & 58.7\std{3.7} & 38.5\std{1.9} & \underline{21.8}\std{13.5} & 45.8\std{6.5} & 54.3\std{4.4} & \textbf{16.3}\std{1.7} & 61.7\std{3.5} & 53.8\std{2.7} \\
\textbf{ZOMP} & \textbf{68.6}\std{0.2} & \textbf{92.8}\std{0.5} & 87.4\std{1.3} & 68.6\std{1.3} & \textbf{85.8}\std{0.1} & 21.3\std{1.0} & \textbf{64.8}\std{1.1} & \textbf{45.4}\std{0.7} & \textbf{30.1}\std{4.5} & \underline{47.8}\std{5.1} & \textbf{57.0}\std{1.8} & 15.6\std{0.7} & \textbf{66.4}\std{0.1} & \textbf{56.9}\std{0.6} \\
\bottomrule
\end{tabular*}

\end{table*}

\begin{figure*}[!htbp]
\centering\includegraphics[width=\textwidth]{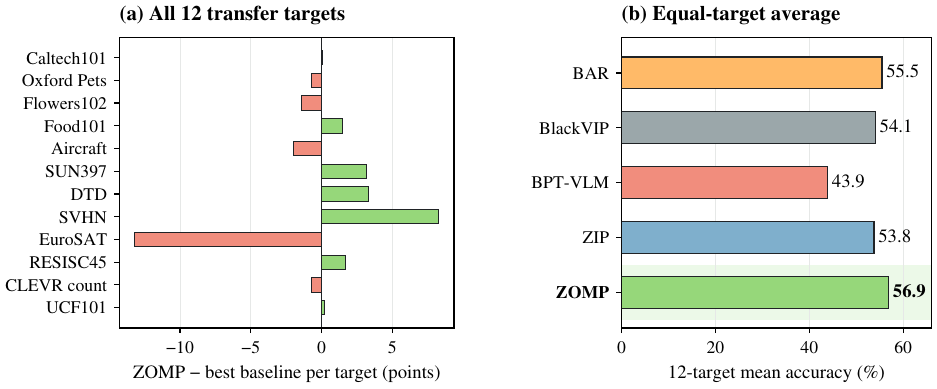}
\caption{\textbf{Cross-dataset transfer against adapted baselines.} (a) ZOMP minus the highest-scoring comparator on each of the 12 targets; the comparator may differ by target. Green denotes a nonnegative margin and coral a negative one. (b) Equal-target mean for each method. All prompts are trained on ImageNet and transferred without updates. Values use the reported one-decimal scores from Table~\ref{tab:cdt}.}
\label{fig:errors}
\end{figure*}

\section{Additional Ablation Details}
\label{app:ablation}
The broad ablation suite contains ten configurations on 13 datasets with three seeds. Table~\ref{tab:broad_components} retains the original component names; Cross-modal LoRA denotes the shared prompt factorization. Its two indented rows separately add scheduling or momentum, and ZOMP combines both. Deep prompt uses plain SPSA. Cross-modal LoRA and its schedule-only variant use clipped SPSA; the momentum-only variant and ZOMP use unclipped SPSA-GC. Thus the original momentum comparison also changes clipping. Comparing momentum-only with ZOMP isolates scheduling under the same GC update.
\begin{table*}[!htbp]
\caption{Original component ablation at 5,000 queries. Each indented row modifies Cross-modal LoRA independently; ZOMP combines rank scheduling and momentum. Unscheduled factorized variants use rank four throughout. Clipping differences are specified in the text. Values are mean\,$\pm$\,SD reconstructed from exact counts.}
\label{tab:broad_components}\label{tab:component_full}
\centering\densetablestyle
\begin{tabular*}{\linewidth}{@{\extracolsep{\fill}}lccccccccccccc>{\columncolor[gray]{0.9}}c@{}}
\toprule
\multicolumn{1}{c}{Method} & \rotatebox[origin=c]{90}{Caltech101} & \rotatebox[origin=c]{90}{OxfordPets} & \rotatebox[origin=c]{90}{Flowers102} & \rotatebox[origin=c]{90}{Food101} & \rotatebox[origin=c]{90}{FGVCAircraft} & \rotatebox[origin=c]{90}{SUN397} & \rotatebox[origin=c]{90}{DTD} & \rotatebox[origin=c]{90}{SVHN} & \rotatebox[origin=c]{90}{EuroSAT} & \rotatebox[origin=c]{90}{RESISC45} & \rotatebox[origin=c]{90}{CLEVR} & \rotatebox[origin=c]{90}{UCF101} & \rotatebox[origin=c]{90}{ImageNet} & \rotatebox[origin=c]{90}{\textbf{Average}} \\
\midrule
Deep prompt & 93.9\std{0.2} & \underline{93.2}\std{0.6} & 73.0\std{3.5} & 86.7\std{0.4} & 16.8\std{8.6} & 67.3\std{0.5} & 54.8\std{2.3} & 47.1\std{1.2} & 74.0\std{4.1} & 71.5\std{0.8} & \underline{33.7}\std{2.3} & 73.0\std{0.9} & 67.6\std{0.8} & 65.6\std{1.5} \\
Cross-modal LoRA & 94.4\std{0.2} & 93.0\std{0.3} & 74.8\std{1.2} & \textbf{87.2}\std{0.1} & \textbf{28.9}\std{0.5} & \underline{68.2}\std{0.1} & \underline{55.8}\std{0.8} & \underline{50.5}\std{1.8} & 75.9\std{0.5} & 71.7\std{0.6} & 31.2\std{1.6} & \underline{73.3}\std{0.5} & \textbf{69.2}\std{0.2} & 67.2\std{0.4} \\
\quad + rank schedule only & 94.4\std{0.4} & 92.9\std{0.2} & 75.1\std{1.9} & \underline{87.1}\std{0.1} & \underline{28.8}\std{0.4} & 68.0\std{0.3} & 55.5\std{1.1} & 47.7\std{2.7} & 77.5\std{4.3} & 71.8\std{0.4} & 28.6\std{3.0} & 72.7\std{0.8} & \textbf{69.2}\std{0.1} & 66.9\std{1.0} \\
\quad + momentum only & \underline{94.7}\std{0.2} & \textbf{93.3}\std{0.3} & \underline{75.4}\std{0.7} & 87.0\std{0.1} & 25.4\std{2.5} & 67.5\std{0.4} & \underline{55.8}\std{1.6} & 49.9\std{5.1} & \underline{77.6}\std{6.5} & \underline{72.3}\std{1.5} & \textbf{35.1}\std{2.1} & 72.3\std{0.8} & 68.6\std{0.1} & \underline{67.3}\std{1.0} \\
\textbf{ZOMP} & \textbf{94.8}\std{0.3} & 93.0\std{0.2} & \textbf{76.5}\std{1.3} & \underline{87.1}\std{0.1} & 28.2\std{0.5} & \textbf{68.4}\std{0.7} & \textbf{56.4}\std{1.3} & \textbf{51.4}\std{6.1} & \textbf{80.4}\std{1.8} & \textbf{73.5}\std{0.5} & \textbf{35.1}\std{2.3} & \textbf{73.6}\std{1.5} & \underline{68.7}\std{0.2} & \textbf{68.2}\std{0.4} \\
\bottomrule
\end{tabular*}

\end{table*}
\begin{table*}[!htbp]
\caption{Schedule and sharing sensitivity across all 13 classification datasets at 5,000 queries. All rows use SPSA-GC; only Momentum + clip adds clipping. Average weights every dataset equally.}
\label{tab:broad_sensitivity}
\centering\densetablestyle
\begin{tabular*}{\linewidth}{@{\extracolsep{\fill}}lccccccccccccc>{\columncolor[gray]{0.9}}c@{}}
\toprule
\multicolumn{1}{c}{Method} & \rotatebox[origin=c]{90}{Caltech101} & \rotatebox[origin=c]{90}{OxfordPets} & \rotatebox[origin=c]{90}{Flowers102} & \rotatebox[origin=c]{90}{Food101} & \rotatebox[origin=c]{90}{FGVCAircraft} & \rotatebox[origin=c]{90}{SUN397} & \rotatebox[origin=c]{90}{DTD} & \rotatebox[origin=c]{90}{SVHN} & \rotatebox[origin=c]{90}{EuroSAT} & \rotatebox[origin=c]{90}{RESISC45} & \rotatebox[origin=c]{90}{CLEVR} & \rotatebox[origin=c]{90}{UCF101} & \rotatebox[origin=c]{90}{ImageNet} & \rotatebox[origin=c]{90}{\textbf{Average}} \\
\midrule
\textbf{ZOMP} & \textbf{94.8}\std{0.3} & \textbf{93.0}\std{0.2} & \textbf{76.5}\std{1.3} & \underline{87.1}\std{0.1} & \textbf{28.2}\std{0.5} & \textbf{68.4}\std{0.7} & 56.4\std{1.3} & 51.4\std{6.1} & \textbf{80.4}\std{1.8} & \underline{73.5}\std{0.5} & \textbf{35.1}\std{2.3} & 73.6\std{1.5} & 68.7\std{0.2} & \textbf{68.2}\std{0.4} \\
Unlock 10\% & 94.4\std{0.3} & \underline{92.9}\std{0.2} & \underline{76.2}\std{1.0} & \textbf{87.2}\std{0.1} & \underline{28.1}\std{0.7} & 67.9\std{0.4} & 56.6\std{1.3} & 48.8\std{2.7} & \underline{79.9}\std{1.8} & 73.0\std{1.8} & 32.6\std{0.8} & \textbf{74.6}\std{1.2} & \textbf{68.9}\std{0.1} & 67.8\std{0.3} \\
Unlock 30\% & \underline{94.6}\std{0.2} & \textbf{93.0}\std{0.2} & 74.3\std{1.7} & \underline{87.1}\std{0.2} & 27.7\std{1.0} & 68.0\std{0.4} & \textbf{57.1}\std{1.3} & 51.2\std{4.0} & 78.3\std{3.9} & \textbf{74.0}\std{0.4} & 33.8\std{2.2} & 73.5\std{1.1} & \underline{68.8}\std{0.2} & 67.8\std{0.4} \\
Schedule $1\to2\to4$ & 94.4\std{0.2} & \underline{92.9}\std{0.4} & 75.2\std{1.3} & 87.0\std{0.0} & 27.1\std{1.2} & 68.0\std{0.3} & 56.5\std{0.5} & \underline{51.7}\std{2.2} & \textbf{80.4}\std{3.4} & 72.7\std{1.5} & \underline{34.9}\std{1.5} & 73.4\std{1.2} & \underline{68.8}\std{0.4} & \underline{67.9}\std{0.1} \\
Momentum + clip & 94.5\std{0.2} & \underline{92.9}\std{0.5} & 75.4\std{0.5} & \textbf{87.2}\std{0.1} & 27.7\std{1.1} & 68.0\std{0.5} & \underline{57.0}\std{1.9} & \textbf{51.8}\std{5.8} & 78.7\std{0.5} & \underline{73.5}\std{1.3} & 33.9\std{1.4} & \underline{73.7}\std{0.6} & \underline{68.8}\std{0.2} & \underline{67.9}\std{0.3} \\
Unshared $U_v,U_t$ & 94.3\std{0.4} & \textbf{93.0}\std{0.3} & 74.0\std{1.1} & 87.0\std{0.2} & 27.7\std{0.3} & \underline{68.2}\std{0.5} & 56.7\std{1.1} & 50.5\std{1.8} & 77.1\std{5.2} & 73.1\std{1.1} & 34.1\std{2.4} & 73.1\std{0.6} & \textbf{68.9}\std{0.2} & 67.5\std{0.5} \\
\bottomrule
\end{tabular*}

\end{table*}
The staged schedule uses rank one through 20\%, rank two through 50\%, then rank four. The unshared variant replaces $U^{(l)}$ with separate branch factors. At full rank this changes optimization coordinates and cross-modal coupling, not the ability to represent arbitrary prompt pairs.

\subsection{Offline Estimator Diagnostic}
\label{app:gradient}
At each saved checkpoint, we freeze a training minibatch of 32 examples, compute an exact reference gradient, and draw 32 independent five-direction SPSA estimates. Each record costs 320 extra loss evaluations and one backward pass, used only for analysis. Adaptation never uses exact gradients.

\begin{figure*}[!htbp]
\centering\includegraphics[width=\textwidth]{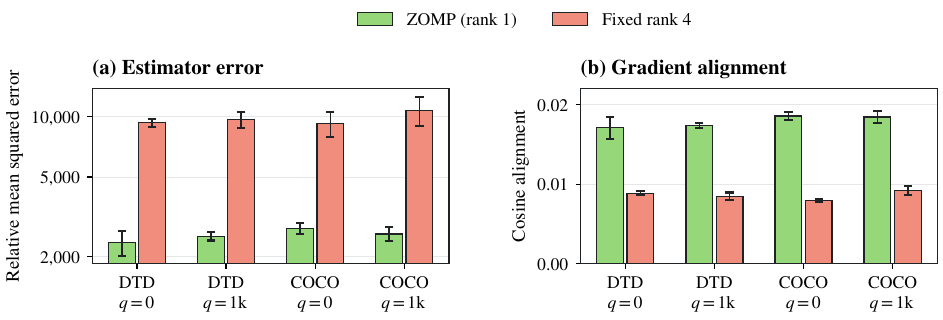}
\caption{\textbf{Early-phase estimator behavior.} Rank-one scheduled search and fixed rank four at matched query checkpoints. Lower relative MSE and higher cosine alignment are favorable. Error bars summarize three seeds, each with 32 estimates; absolute alignments remain small. The MSE axis is logarithmic.}
\label{fig:gradient}
\end{figure*}

There are 36 records: DTD and COCO, three parameterizations (scheduled, fixed rank four, direct), checkpoints $q=0,1{,}000$, and three seeds. Both scheduled checkpoints remain at rank one because $q/Q=0.2$ is still in the restricted phase. Relative MSE normalizes squared estimator error by squared reference-gradient norm; raw variance alone is not comparable across parameterizations. Figure~\ref{fig:gradient} concerns the early phase and does not measure a post-activation variance change.

\section{Resources and Reproducibility}
\label{app:resources}
\begin{figure*}[!htbp]
\centering\includegraphics[width=\textwidth]{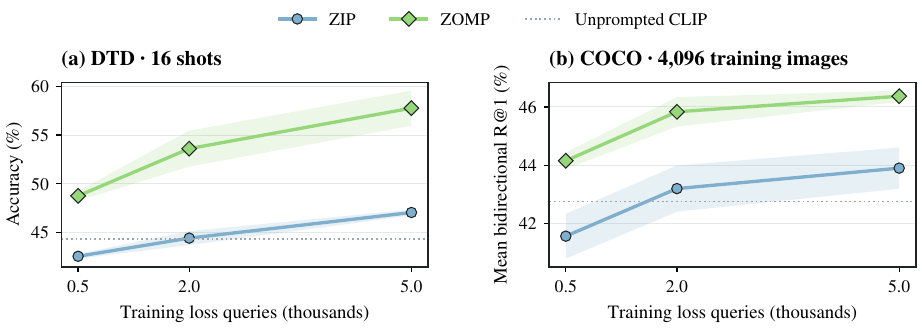}
\caption{\textbf{Independent query-budget runs.} Held-out performance after 500, 2,000, or 5,000 training-loss queries. Each point has its own training horizon; lines connect means and shaded bands show sample SD over three seeds. These are query counts, whereas COCO's 4,096 images specify its training-set size.}
\label{fig:budgets}
\end{figure*}
\paragraph{Query accounting.}
A query is one scalar minibatch loss evaluation, not one image, epoch, or optimizer step. ZOMP normally uses ten queries per update; BAR uses a pivot plus up to five probes. BPT candidate fitness can consume multiple minibatch queries and stops before an incomplete candidate. Thus the common cap need not yield an identical number of updates or exactly the same final consumed count. Validation, test inference, development, and offline gradient analysis are tracked separately from the main training cap.

Related work on test-time scaling in reasoning LLMs emphasizes protocol-specific compute accounting and reproducibility~\citep{hariri2026testtime}. Here, the query budget governs prompt adaptation; deployment uses fixed prompts and incurs a separate inference cost. Reporting these stages separately makes the resource comparison interpretable.

\paragraph{Memory and time.} Table~\ref{tab:memory} summarizes the COCO resource comparison.
Experiments run on NVIDIA RTX A6000 GPUs. Peak allocated training memory differs from reserved or total device/process memory. On ViT-L/14 with batch size 32, ZOMP uses 1.96\,GiB and AdamW 15.30\,GiB. Different batch sizes prevent a fixed-batch scaling comparison across backbones. Concurrent runs share devices, so their elapsed times are not used to claim a hardware speedup.

\paragraph{Reproduction materials.}
The five-dataset external retrieval comparison contains 150 trained models. The selected diagnostic comparison evaluates these same models on five benchmarks, giving 750 evaluations plus five unprompted references. Per-run metrics, seed aggregates, checkpoint identities, and source hashes accompany the manuscript. Tables and editable vector figures regenerate from these numerical records without loading model weights. The source package includes the official ACL style unchanged. Dataset images remain subject to their original access conditions.

\begin{figure*}[!htbp]
\centering\includegraphics[width=\textwidth]{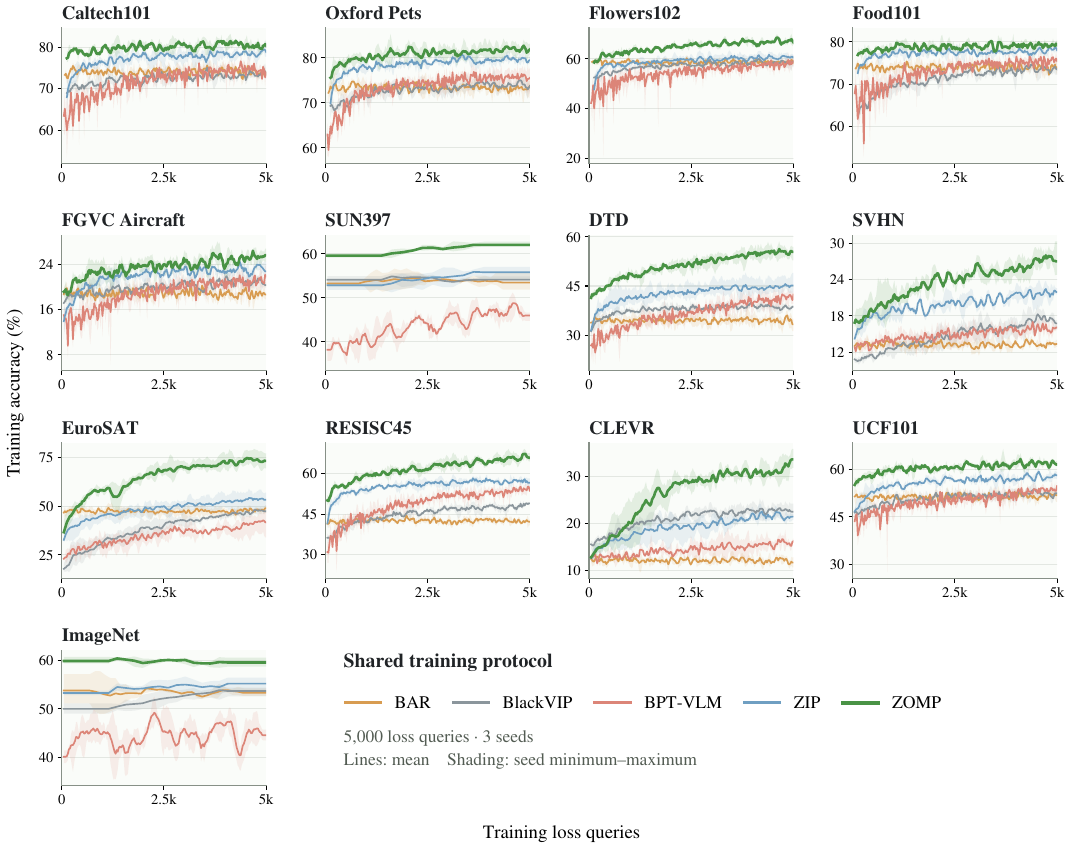}
\caption{\textbf{Training-accuracy curves across all 13 classification datasets.} Lines show the original smoothed three-seed means; bands span the seed minimum and maximum. All methods use a 5,000-query training horizon.}
\label{fig:original_train_curves}
\end{figure*}

\paragraph{Classification training curves and efficiency.}
\label{app:original_efficiency}
Figure~\ref{fig:original_train_curves} reports the original training-accuracy curves across all 13 classification datasets. Lines show three-seed means and shaded bands span the seed minimum and maximum. The plotting routine smooths ten logged observations and interpolates onto a common query grid, with logged positions scaled to the 5,000-query horizon. These curves describe training behavior; Figure~\ref{fig:budgets} reports held-out performance from independent budget runs.

Table~\ref{tab:original_efficiency} retains the original efficiency comparison. Its Queries column averages the first crossing of a per-dataset training-accuracy target: the smallest peak among the compared methods over the full horizon. The training cap remains 5,000 queries. Profiling uses a single RTX A6000 and ViT-B/16, with batch size one for VRAM, inference latency, throughput, and FLOPs. These quantities are averaged over the 13 datasets; latency and FPS are averaged separately. Train reports mean elapsed training time over the 13 datasets and three seeds. The profiling settings differ from the batch-128 COCO training-memory measurements in Table~\ref{tab:memory}.

\begin{table*}[!htbp]
\caption{Original classification efficiency comparison, averaged over 13 datasets. Queries measures training-target attainment as defined in the text. \textbf{Best} and \underline{second-best} per column; lower is better except FPS and Acc.}
\label{tab:original_efficiency}
\centering\tablestyle
\begin{tabular*}{\linewidth}{@{\extracolsep{\fill}}lcccccccc@{}}
\toprule
\multicolumn{1}{c}{Method} & \shortstack{Params\\(k)} & Queries & \shortstack{VRAM\\(GB)} & \shortstack{Train\\(min)} & \shortstack{Infer\\(ms/img)} & FPS & GFLOP & \shortstack{Acc\\(\%)} \\
\midrule
BAR & 37.6 & 1821 & \textbf{0.36} & \underline{63.2} & \underline{43.0} & \underline{23.8} & \textbf{17.6} & 56.2 \\
\textsc{BlackVIP} & 9.9 & 2442 & \underline{0.77} & 86.5 & \textbf{36.9} & \textbf{27.2} & \underline{35.2} & 58.9 \\
BPT-VLM & \underline{4.0} & 2492 & 1.00 & 64.0 & 396.4 & 7.5 & 490.5 & 57.5 \\
\textsc{ZIP} & \textbf{0.4} & \underline{638} & 0.95 & 71.7 & 487.8 & 3.7 & 490.1 & \underline{62.2} \\
\textbf{ZOMP} & 46.2 & \textbf{191} & 0.96 & \textbf{57.7} & 380.1 & 7.9 & 490.5 & \textbf{68.2} \\
\bottomrule
\end{tabular*}

\end{table*}

\end{document}